\documentclass[5p,times]{elsarticle}
\usepackage{booktabs}

\usepackage{graphicx}
\usepackage{schemata}
\usepackage{color,soul}
\usepackage{amsmath}
\usepackage{multirow}
\usepackage{amsfonts}
\usepackage[table]{xcolor}% http://ctan.org/pkg/xcolor. colours cells
\DeclareMathSymbol{*}{\mathbin}{symbols}{"03} % \ast
\DeclareMathSymbol{\ast}{\mathbin}{symbols}{"03}
\usepackage{comment}

\usepackage{makecell}

\usepackage{soul}
\usepackage{units}
\usepackage{tikz}
\usepackage{standalone, times}
\usetikzlibrary{mindmap,backgrounds}
\usetikzlibrary{positioning}

\usepackage{multirow}
\usepackage{algorithm}
\usepackage{algorithmic}

\usepackage{enumitem}
\usepackage{float}

\usepackage[toc,page]{appendix}
\usepackage[modulo]{lineno}
\usepackage{graphicx}
\usepackage{varioref}
\usepackage{array}
\usepackage{amsmath}
\usepackage{booktabs}
\usepackage{multirow}
\usepackage{subcaption}
\usepackage{array,arydshln}
\usepackage{amssymb}
\usepackage{hhline}
\usepackage{times,rotating}
\usepackage{comment}
\usepackage{tikz}
 
 \tikzset{variable/.default=}
\usepackage{tabularx,ragged2e}
\newcolumntype{T}[1]{>{\raggedright\arraybackslash}p{#1}}
\newcolumntype{M}[1]{>{\centering\arraybackslash}m{#1}}

\newcolumntype{L}[1]{>{\raggedright\let\newline\\\arraybackslash\hspace{0pt}}m{#1}}
\newcolumntype{C}[1]{>{\centering\let\newline\\\arraybackslash\hspace{0pt}}m{#1}}
\newcolumntype{R}[1]{>{\raggedleft\let\newline\\\arraybackslash\hspace{0pt}}m{#1}}

\usepackage{tikz}
\usepackage[edges]{forest}
\useforestlibrary{linguistics}
\forestapplylibrarydefaults{linguistics}
\usepackage{pifont}

\usepackage{framed} % Framing content
\usepackage{multicol} % Multiple columns environment
\usepackage{nomencl} % Nomenclature package
\usepackage{hologo}
\usepackage{booktabs}       % professional-quality tables
\usepackage{amsfonts}       % blackboard math symbols
\usepackage{nicefrac}       % compact symbols for 1/2, etc.
\usepackage{microtype}      % microtypography
\usepackage{diagbox} %backslashbox uses it for a diagonally split table cell
\usepackage{dsfont}
\usepackage{bm} % math bold. Example of use: %$\bm{BWT^{+}}$
\usepackage{amsmath}
\usepackage{graphicx}
\usepackage{amsfonts}
\usepackage{subcaption}
\usepackage{color, colortbl}
\usepackage{breakcites}
\usepackage{pdfpages}
\usepackage{rotating}
\usepackage{float}
\usepackage{dirtree} %tree drawing
\usepackage{afterpage}
\usepackage{longtable}
\usepackage[colorinlistoftodos]{todonotes} % B
\usepackage[colorlinks=true, allcolors=blue]{hyperref}

\definecolor{orange}{HTML}{FFC17D}
\definecolor{green}{HTML}{A1D68B}
\definecolor{lightgray}{HTML}{E8E8E8}

\usepackage{hyperref} % redefinition needed to work with todo notes

\makenomenclature
\renewcommand*\nompreamble{\begin{multicols}{2}}
\renewcommand*\nompostamble{\end{multicols}}

\usetikzlibrary{arrows.meta,shadows.blur}

\begin{document}

\begin{frontmatter}

\title{A Practical guide on Explainable AI Techniques applied on Biomedical use case applications}

\author[a,b,c]{Adrien Bennetot}
\ead{adrien.bennetot@ensta-paris.fr}
\author[d]{Ivan Donadello}
\ead{ivan.donadello@unibz.it}
\author[c,f]{Ayoub El Qadi}
\ead{ayoub.el_qadi_el_haouari@etu.sorbonne-universite.fr}
\author[e]{Mauro Dragoni}
\ead{dragoni@fbk.eu}
\author[f]{Thomas Frossard}
\ead{tfd@tinubu.com}
\author[h]{Benedikt Wagner}
\ead{Benedikt.Wagner@city.ac.uk}
\author[i]{Anna Saranti}
\ead{anna.saranti@medunigraz.at}
\author[k,c]{Silvia Tulli}
\ead{silvia.tulli@gaips.inesc-id.pt}
\author[g]{Maria Trocan}
\ead{maria.trocan@isep.fr}
\author[c]{Raja Chatila}
\ead{raja.chatila@sorbonne-universite.fr}
\author[i,j]{Andreas Holzinger}
\ead{andreas.holzinger@medunigraz.at}
\author[h]{Artur d'Avila Garcez}
\ead{a.garcez@city.ac.uk}
\author[a,l]{Natalia D\'iaz-Rodr\'iguez}
\ead{ndiaz@decsai.ugr.es}

\address[a]{ENSTA, Institut Polytechnique Paris and INRIA Flowers Team, Palaiseau, France}
\address[b]{Segula Technologies, Parc d'activit\'e de Pissaloup, Trappes, France}
\address[c]{Sorbonne Universit\'e, Paris, France}
\address[d]{Free University of Bozen-Bolzano, Italy}
\address[e]{Fondazione Bruno Kessler, Trento, Italy}
%\address[e]{DaSCI Andalusian Institute for Data Science, University of Granada, Spain.}
\address[f]{Tinubu Square, %169 Quai de Stalingrad, 92130 
Issy-les-Moulineaux, France.}
\address[g]{Institut Supérieur d'Électronique de Paris (ISEP), Issy-les-Moulineaux, France.}
\address[h]{City University London, U.K. %Department of Computer Science
}
\address[i]{Medical University Graz, Austria}
\address[j]{Alberta Machine Intelligence Institute, Edmonton, Canada}
\address[k]{INESC-ID and Instituto Superior T\'ecnico, Lisbon, Portugal}
\address[l]{% Dpt. of Computer Science and Artificial Intelligence, 
Andalusian Research Institute in Data Science and Computational Intelligence (DaSCI), University of Granada, Spain}

\newpage

\begin{abstract} 
Last years have been characterized by an upsurge of opaque automatic decision support systems, such as Deep Neural Networks (DNNs). Although they have great generalization and prediction skills, their functioning does not allow obtaining detailed explanations of their behaviour. As opaque machine learning models are increasingly being employed to make important predictions in critical environments, the danger is to create and use decisions that are not justifiable or legitimate. Therefore, there is a general agreement on the importance of endowing machine learning models with explainability. EXplainable Artificial Intelligence (XAI) techniques can serve to verify and certify model outputs and enhance them with desirable notions such as trustworthiness, accountability, transparency and fairness. This guide is meant to be the go-to handbook for any audience with a computer science background aiming at getting intuitive insights on machine learning models, accompanied with straight, fast, and intuitive explanations out of the box. This article aims to fill the lack of compelling XAI guide by applying XAI techniques in their particular day-to-day models, datasets and use-cases. Figure \ref{fig:Flowchart} acts as a flowchart/map for the reader and should help him to find the ideal method to use according to his type of data. In each chapter, the reader will find a description of the proposed method as well as an example of use on a Biomedical application and a Python notebook. It can be easily modified in order to be applied to specific applications. 

\end{abstract}


\end{frontmatter}

\section{Introduction}

\begin{figure*}[]%htbp!]
\centering
\includegraphics[scale=0.50]{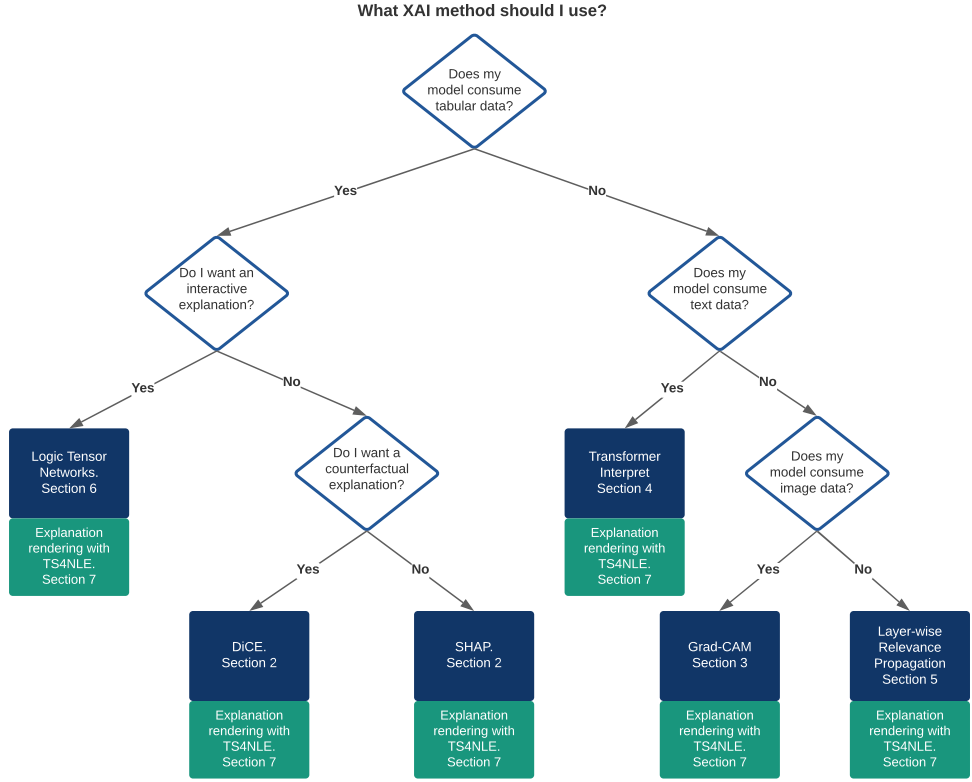} 
\caption{Map/Flowchart of the XAI methods described in the article. It is not meant to be exhaustive regarding all the existing XAI methods as it presents only the existing methods in the paper. Moreover, the Layer-wise relevance propagation method is recommended when the data are neither textual, tabular nor images, but it can be used on these type of data as well. Also, SHAP can be used on images but this particular use-case is not in the scope of this article.}
\label{fig:Flowchart}
\end{figure*}

In systems based on machine learning, one has to consider that there will unavoidably be faulty system decisions. This is due to several reasons, originating from data (e.g., bias) or from the system's design (e.g., network structure, connectivity, optimization process, code quality management, etc). It leads to a given system performance in terms of accuracy, false negatives or false positives compared to ground truth. This is actually an unavoidable feature of DNNs, and not simply bugs in the system that can be corrected.  This performance can be largely insufficient to trust the system and could lead to catastrophic outcomes in critical applications e.g., where human lives can be at stake. An interesting approach is to consider this situation from a software engineering viewpoint. Following principles of dependable and robust system design could insure a safer system operation. Therefore, there is a need for tools for understanding the system's behavior and outputs, i.e. the need for explainability. It refers to the details and reasons a model gives to make its functioning clear or easy to understand for the receiver of the explanation \cite{arrieta2020explainable}.  

%However this is based on proper tools for understanding the system's behavior and outputs among which explainability..

%Explainability 
%The literature makes a clear distinction among models that are interpretable by design and those that can be explained by means of external methods. 

EXplainable AI (XAI) techniques are increasingly being used by a wider audience and are starting to be applied in multiple fields in industry and in academia. More and more techniques appear and it is often complex to interpret or convert their explanatory elements into an actionable explanation, i.e., that either a developer or an expert can transform into an action to fix the model. We present a guide of some of the most used XAI methods producing explanations in common formats applied on models ingesting different type of data (image, tabular, textual and graphs) and models using neural-symbolic computation. We also consider some user interface aspects to better interact with non technical users, such as neural-symbolic computation models producing natural language explanations and counterfactual explanations. 

As black-box Machine Learning models are increasingly being employed to make important predictions in critical contexts \cite{Gianotti21}, the demand for transparency is increasing from the various stakeholders in AI \cite{Preece18Stakeholders}. This is particularly true in the field of healthcare, where operations need to be trusted for ethical and safety reasons \cite{Pawar2020}. With this motivation, all techniques presented are accompanied by an example of application related to the biomedical field, such as medical image recognition or bipolar disease prediction. 

%The goal of this guide is to give a practical overview of some common XAI techniques with hands-on material. Data, models and Python \textit{Google Colab} interactive notebooks are free and open source to be easily reused, adapted or taught and are presented with a systematic introduction of fundamental theories and common practices with practical use case and suitability analysis for any application, task or data type, with concrete examples of use cases and interactive codes.

The goal of this guide is to serve as a practical and rapidly usable tool for any developer wishing to obtain explanatory elements on the behaviour of his Deep Learning model. These can be used within and in complement with other implementations to regulate, audit and govern AI systems \cite{de2021companies}. Contrarily to other guides that focus on some particular data type \cite{Camburu2020}, we provide a broader view on XAI methods by addressing most of the data types and issues faced by users wishing to explain their Deep Learning models.

%\ivan{Now we should say something about XAI for biomedical/healthcare domain. Otherwise it seems just a random resubmission. I would suggest to have a look at the abstract and introduction of \url{https://www.researchgate.net/publication/342600571_Explainable_AI_in_Healthcare} and \url{https://www.mdpi.com/2673-7426/2/1/1/htm}}
To this end, we describe an XAI method for each of the most common data types: tabular in Section \ref{tab}, images in Section \ref{images} and textual in Section \ref{text} as well as a general method in Section \ref{LRP}. We accompany these techniques with a description of the neuro-symbolic methods used to make interactive explanations in Section \ref{Nesy} as well as a method to render XAI Explanations through Natural Language Generation in Section \ref{NLG}. Data, models and Python \textit{Google Colab} interactive notebooks are free and open source to be easily reused, adapted or taught. They are presented with a systematic introduction of fundamental theories and common practices with use cases and suitability analysis for any application, task or data type, with concrete examples and interactive codes. Table \ref{Summary} references the different methods presented in this article and their associated working demonstrations (i.e., \textit{Google Colab} notebooks) to facilitate the reader exploration.

\begin{table*}
\centering
\begin{tabular}{lllll}
\hline
\multicolumn{1}{c}{\textbf{XAI Method}}                                     & \textbf{Data Type}                                               & \textbf{Explanation Type} & \textbf{ML Task Explained} & \textbf{Dataset used} \\ \hline
SHAP \cite{lundberg2017unified}                                                                         & Tabular                                                          & Feature Importance         & Bipolar Disease Prediction          & Simula  Depresjon \cite{garcia2018depresjon}\textsuperscript{1}  
\\
\multicolumn{5}{l}{Original \textbf{SHAP} Repository: \tiny{\url{https://github.com/slundberg/shap}}}                                                                                                                                                                            \\
\multicolumn{5}{l}{SHAP practical case guide: \tiny{\url{https://colab.research.google.com/drive/1AxdhD-ZkZya57-ePk6Nqg0Z8P2eMu9XX?usp=sharing}}
}                                                                                                                                                                       \\ \hline
DiCE \cite{Mothilal_2020}                                                                          & Tabular                                                          & Counterfactual            & Bipolar Disease Prediction         & Simula  Depresjon \cite{garcia2018depresjon}              \\
\multicolumn{5}{l}{Original \textbf{DiCE} Repository: \tiny{\url{https://github.com/interpretml/DiCE}} 
}                                                                                                                                                                            \\
\multicolumn{5}{l}{DiCE practical case guide: \tiny{\url{https://colab.research.google.com/drive/12jw91RouPBc9slFwB2OWRwYB6Ckviiv3?usp=sharing}}
}                                                                                                                                                                         \\ \hline
Transformers Interpret (TI) \cite{Pierse_Transformers_Interpret_2021}                                                       & Textual                                                          & Feature Importance        & Sentiment Analysis          & MultiNLI corpus %Multi-Genre Natural Language Inference (MultiNLI) corpus 
\cite{williams2017broad}\textsuperscript{2}              \\
\multicolumn{5}{l}{Original \textbf{TI} Repository: \tiny{\url{https://github.com/cdpierse/transformers-interpret}}}                                                                                                                                                                         \\
\multicolumn{5}{l}{TI practical case guide: 
\tiny{\url{https://colab.research.google.com/drive/1XGGXUYNC1M_jlmQUV5dZB3HVdQRXeghd}}
}                                                                                                                                                                         \\ \hline
Grad-CAM \cite{selvaraju2016grad}                                                                   & Image                                                            & Visual                    & Image Classification          & TCGA and Target datasets \cite{grossman2016toward}\textsuperscript{3}  (GDC)               \\
\multicolumn{5}{l}{Original \textbf{Grad-CAM} Repository: \tiny{\url{https://keras.io/examples/vision/grad_cam/}}
}                                                                                                                                                                         \\
\multicolumn{5}{l}{Grad-CAM practical case guide: \tiny{\url{https://colab.research.google.com/drive/1ZXznvG_G1Y-JyHX9a_x6yKrXHhMp6tpm\#scrollTo=gSx4Fef2JlN-}}
}                                                                                                                                                                         \\ \hline
\begin{tabular}[c]{@{}l@{}}Layer-wise Relevance \\ Propagation (LRP) \cite{montavon2017explaining}\end{tabular} & \begin{tabular}[c]{@{}l@{}}Graph\end{tabular} & Visual                    & Image Classification          & TCGA and Target datasets \cite{grossman2016toward}

\\
\multicolumn{5}{l}{Original \textbf{LRP for Graphs} Repository: \tiny{\url{https://git.tu-berlin.de/thomas_schnake/demo_gnn_lrp}}
}                                                                                                                                                                         \\
\multicolumn{5}{l}{LRP for Graphs practical case guide: \tiny{\url{https://colab.research.google.com/drive/166FYIwxblfrEltkYqY_jiJoAm9VLMweJ?usp=sharing}}
}                                                                                                                                                                         \\ \hline
\begin{tabular}[c]{@{}l@{}}Logic Tensor \\ Networks (LTN) \cite{Serafini2016}\end{tabular}            & Textual                                                          & Interactive               & Violence Risk Prediction          & COMPAS  ProPublica dataset \cite{larson2016data}\textsuperscript{4}          \\
\multicolumn{5}{l}{Original \textbf{LTN} Repository: \tiny{\url{https://github.com/logictensornetworks/logictensornetworks}}
}                            

\\
\multicolumn{5}{l}{LTN practical case guide: \tiny{\url{https://colab.research.google.com/drive/1Ip9Yb9gVRSRqaBKY9gOpiWn9pq3LovWG?usp=sharing}}
}                                                                                                                                                                         \\ \hline
TS4NLE \cite{Donadello19}                                                                       & Textual                                                          & Natural Language          & Explanation Rendering   & Unified Medical Language System %(UMLS)
\cite{bodenreider2004unified}\textsuperscript{5}           \\
\multicolumn{5}{l}{Original \textbf{TS4NLE} Repository: \tiny{\url{https://github.com/ivanDonadello/TS4NLE}} 
}                                                                                                                                                                   \\
\multicolumn{5}{l}{TS4NLE practical case guide: \tiny{{\url{https://colab.research.google.com/drive/1iCVSt7TFMruSzeg5DswLOzOR1n7xATbz}}} 
}                                                                                                                                                                         \\ \hline
\end{tabular}

\afterpage{\afterpage{\footnotetext{\textsuperscript{1}Depresjon dataset \url{https://datasets.simula.no/depresjon/}}
\footnotetext{\textsuperscript{2}\url{https://huggingface.co/datasets/multi_nli}}
\footnotetext{\textsuperscript{3}Use Data Transfer Tool in %\url{https://docs.gdc.cancer.gov/Data\_Transfer\_Tool/Users\_Guide/Getting\_Started/#downloading-the-gdc-data-transfer-tool}
%\url{https://docs.gdc.cancer.gov/Data#_Transfer#_Tool/Users#_Guide/Getting#_Started/#downloading-the-gdc-data-transfer-tool}
\url{https://docs.gdc.cancer.gov/Data_Transfer_Tool/Users_Guide/Getting_Started/\#downloading-the-gdc-data-transfer-tool}
to download TCGA (Cancer Genome Atlas) and the TARGET (Therapeutically Applicable Research to Generate Effective Treatments) program datasets \cite{grossman2016toward}, available at the GDC Data Portal \url{https://portal.gdc.cancer.gov/} and Legacy Archive \url{https://gdc.cancer.gov/} through \url{https://gdc.cancer.gov/access-data}.}%
%\url{https://portal.gdc.cancer.gov/}}
\footnotetext{\textsuperscript{4}{\url{https://github.com/propublica/compas-analysis}}}

\footnotetext{\textsuperscript{5}Unified Medical Language System (UMLS) \url{https://www.nlm.nih.gov/research/umls/index.html}}
}}

\caption{Different XAI methods exposed in the paper as well as the link to the associated \textit{Google Colab}, which can be used to apply the method to his own use-case. Note that certain methods can also be used for other data types than the one described in the table but only the data types treated in the guides are mentioned. All guides are accessible in \url{https://github.com/NataliaDiaz/XAI-guide}.}

\label{Summary}
\end{table*}

\section{XAI techniques for tabular data} \label{tab}% SHAP (Ayoub)}
In this section we will explore two different post-hoc explanations methods: SHapley Additive exPlanations (SHAP) and Diverse Counterfactual Explanations.
\subsection{SHAP Analysis}
Model-agnostic interpretation methods separate the explanations from the machine learning model. The flexibility in such methods lies in the ability to use the method to explain any machine learning model. %\cite{ribeiro2016modelagnostic}. 
In this section, we will focus on one of most used model-agnostic method, the SHapley Additive exPlanations \cite{lundberg2017unified} and its use for tabular data. 

SHapley Additive exPlanations (SHAP) %\cite{lundberg2017unified} 
is based on game theory, in particular on the SHAPley values. The original framework was built in order to re-allocate in a rightful way the gain of a cooperative game among players. SHAP decomposes the prediction of a model among all features involved by using an additive feature attribution analysis:
%%%%%Corrected 
%\nat{say here what is g called and what it means to introduce it}
\begin{equation}
\centering
  g(x^{'}) = \phi_0 + \sum_{i=1}^{M}\phi_i x_{i}^{'}
\end{equation}

where $g(x^{'})$ is the explanation model that matches the original model $f(x)$ when $x=h_x(x')$ and where $x^{'} \in \{0, 1\}^{M}$, $M$ is the number of input features and $\phi_i \in \mathbb{R}$. $\phi_0$ represents the baseline model (i.e., the model without the feature $i$) while $\phi_i$ corresponds to the contribution of feature $i$ to the model prediction:

\begin{equation}
    \phi_i = \sum_{S\subseteq N \setminus i} \frac{\mid S\mid ! (M -  |S| - 1)!}{M!} [f_{X}(S \cup {i}) - f_{X}(S)]
\end{equation}
where $N$ is the set of all input features.
The inner functioning of SHAP considers, for each feature $i$, two different models: $f_{S\cup \{i\} }(x)$ and $f_S(x)$. Then it computes the difference in prediction between both models. This difference is attributed to feature $i$.

Since it is computationally expensive to consider all possible sets of features $S$ and average the difference in prediction due to the feature $i$, SHAP generates a random sampling of the possible sets of $S$ to compute the average. This average represents the estimated feature importance. SHAP exhibits numerous desirable properties, such as singularity detection (i.e., if the feature is locally zero, the SHAP value is zero), local accuracy (i.e., for a specific input x, the explanation model matches the output of the model $f$ for the simplified input $x'$)
%\nat{unclear, explain better local accuracy}
and consistency (i.e., if in a second model approximation with a different subset of features the contribution of the feature is higher, so will be its SHAP value).

\subsubsection{SHAP Use case: Predicting Bipolar Disease} \label{use case shap}
%Corrected
%\nat{is this definition 100\% correct? doesnt sound to me, check with dictionary}

Bipolar disorder, formerly called manic depression, is a mental health condition that causes extreme mood swings that include emotional highs (mania or hypomania) and lows (depression) which often come accompanied by different features (i.e., physical and psychological features). For this particular use case, we build a model based on Extreme Gradient Boosting (XGBoost). Then, we apply SHAP in order to help psychiatrists understand the causes behind a potential patient tendency towards a mania or depression episode.

\subsubsection{SHAP explanation visualization}
%The SHAP technique facilitates the understanding of the model by displaying what features have been the most relevant for the model and their impact in the final prediction. On one hand, the SHAP analysis show us what  variables  influenced model's output the most. On the other hand, SHAP analysis does not explain how the magnitude of the different features affects the output of the model. This is mainly due the following reasons: highy imbalanced, considerable volume of missing data and the scarce features used.

The SHAP technique facilitates the understanding of the model by displaying what features had the most impact, i.e., 
contributed the most to the model prediction. It also reveals how the magnitude of the different features affects --positively or negatively-- to the probability of suffering a bipolar disease in the future.

%To add the refernce to the papaer where you can find the obtained results
%\begin{figure}[htbp!]
%\centering
%\includegraphics[scale=0.3]{figures/SHAP_values.png} 
%\caption{Contribution of each explanatory feature to the final prediction based on SHAPley analysis of contribution decomposition for the default prediction. Axis x represents whether the value of the feature contributes positively (negative values in x axis reduce the probability of default and viceversa).While some features contribution are easy to interpret, since high or low values are homogenized and concentrated in one range of the horizontal plot in a single colour, others are harder. When this is not the case, it means we can not conclude how high (or low) values of these variables affects the probability of default when talking about a high (or low) value of such feature in a general manner. Features are sorted according to the their relevance (i.e., SHAP average absolute value). Source: \cite{qadi2021explaining}}
%\label{fig:SHAP}
%\end{figure}

\begin{figure}[htbp!]
\centering
\includegraphics[scale=0.45]{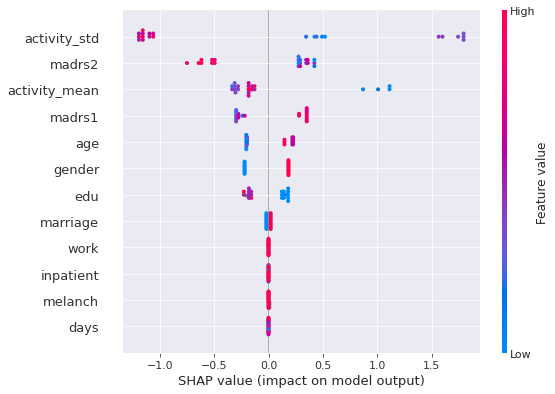}
\caption{%SHAP values for each feature. 
X-axis represents the SHAP value (i.e., contribution) for each feature of the bipolar disease prediction model, while y-axis indicates the features, ranked by importance from top to bottom. Each point represents a data point (i.e., a single patient). Feature values are encoded using a color gradient. Negative values in x-axis correspond to an increase in the probability of  being diagnosed with a bipolar disease.}
%%%%%AEI: Explained Below
%\cnat{madrs1 y madrs2 what is the difference ?and why its shap interpretation is the opposite in shap plot in fig 2?}}
\label{fig:SHAP}
\end{figure}

%Contribution of each explanatory feature to the final prediction based on SHAPley analysis of contribution decomposition for the default prediction. X axis represents the feature contribution value (negative values in x axis increase  the probability of default and viceversa). Features are sorted according to the their relevance (i.e., SHAP average absolute value)
In the studied case Fig. \ref{fig:SHAP} shows at the top the most relevant features for the model when predicting a bipolar disease: the variation in the \textit{activity} measurement,  
the Montgomery Asberg Depression Rating Scale (MADRS) at the moment the Actigraph was stopped (\textit{madrs2}), the mean of the \textit{activity} measurement and the MADRS at the moment the actigraph was initiated (\textit{madrs1}). %\cnat{Ayoub explain what is madrs1 here and in the table's caption where it is used}
The analysis results are intuitive and show that high values in the variation of the activity measurements (\textit{activity\_std}) decrease the probability of being diagnosed with a bipolar disease\footnote{%Data and models are 
SHAP guide online: \url{https://colab.research.google.com/drive/1AxdhD-ZkZya57-ePk6Nqg0Z8P2eMu9XX?usp=sharing} adapted from \url{https://github.com/slundberg/shap}}.

\subsubsection{SHAP values suitability analysis: pros and cons}
Among additive feature attribution methods, SHAP is the only possible consistent, locally accurate method that obeys the missingness property (i.e., a missing feature gets an attribution of zero). However it is computationally expensive since as the number of features increases, the number of possible combinations combinatorially explodes, leading to an expensive computation time. For tree-based models, there is a version of SHAP \cite{lundberg2019consistent} that allows to compute the exact SHAP values faster, in polynomial time, by keeping track of the number of subsets S that flow into each node of the tree. Another problem is that the SHAPley value can change with the order of features selected, and thus, for an exact computation of SHAP values, all possible combinations of subsets must be considered. %However, in practice, this approximation suffices?

% ToDo move to Conclusions? 
%Future works in this domain 
Generally, XAI techniques such as SHAP only focus on explaining the model's inner functioning. However, they do not compare the level of alignment of the ML model explanation with human expert interpretations (i.e., psychiatrists). Hence, there is a need for ML-models in this domain to meet experts criteria in order to allow trust. This is a crucial requirement for ML model adoption in % applications involving critical decisions.
critical decision making.

\subsection{DiCE: Diverse Counterfactual Explanations}
Some XAI techniques for tabular data focus on explaining the model by measuring the features that impacted the most the prediction. On the other hand, there are methods that explain the model by providing information of feature-perturbed versions of the analyzed instance. These methods fall into the counterfactual explanations methods. In this section we focus on Diverse Counterfactual Explanations (DiCE) \cite{Mothilal_2020} and show how to implement it for a critical use case: Bipolar Disease Prediction.

DiCE considers the problem of generating counterfactual explanations from a set of counterfactual (CF), i.e., alternative events to a given model output.
%%%%%%%Corrected: It is used by the authors in the original paper
%\nat{CE? where the F comes from, is this accronym used elsewere?: }
This is set as an optimization problem. Ideally the set of CF examples should balance the variety of the suggested CF instances (diversity) with the capability of the stakeholder to meet the suggested changes (proximity) proposed by the CF framework. Furthermore, the CF explanations need to be aligned with human experts' criteria. 

We present term by term the elements of the function to minimize \cite{Mothilal_2020}. The first term to be encoded mathematically is the concept of diversity. Diversity is captured building on determinantal point processes (DPP), a method for solving the subset selection problem with diversity constraints.
%%%%%%%%%%Corrected: The underscore is how they present this term in the original paper

%\nat{the underscore should not show, should be a sub index, same fin eq 5 for diversity. determinant? is not said what it stand for (det), say every term always with words what it means, make sure this is done in every equation.  }%, and use mathcal command for L of loss , what ys yloss }

\begin{equation}
    dpp\_diversity  = det(K) 
\end{equation}
where  $det$ is the computation of the determinant of matrix $K$ with  $K_{i,j} = \frac{1}{1+dist(c_i,c_j)} $ and $dist(c_i,c_j)$ is a distance metric. $c_i$ represents each generated counterfactual explanation.

Proximity is quantified as the negative distance between the CF example's features and the original input's. For each generated CF ($c_i$) we compute the distance between the CF and  the input $x$ which is the d
\begin{equation}
Proximity  := -\frac{1}{k} \sum_{i=1}^{k} dist(c_i, x) 
\end{equation}
%\nat{where ci is? and x is? say it better not here but in the phrase above to save space e.g. example's features x and the original input x - like this in every formula, no term left unexplained}
$C$ is the generated set of $k$ CFs generated for example $x$ that minimizes the following function (as in \cite{Mothilal_2020}):
\begin{gather}
C(x)  = arg\;min_{c_1, c_2, ..., c_k} \frac{1}{k}\sum_{i=1}^{k} yloss(f(c_i), y) \\
+ \frac{\lambda_1}{k} \sum_{i=1}^{k} dist(c_i, x) \nonumber
- \lambda_2 dpp\_diversity(c_1, c_2, ..., c_k)
\label{eq: dice equation}
\end{gather}
where $yloss()$ measures the distance between the output of the model for the CF generated  $f(c_i)$ and the output we desire, i.e. the  generated CF example. $C$ is the set of $k$ CFs that are close to example $x$, with a high diversity within the CF generated and for which the outcome of the model is as close as possible to the desired class. Both $\lambda_1$ and $\lambda_2$ are hyperparameters that balance the three parts of the loss function.
 
%%%%%%%Corrected
%\nat{a metric is used for reporting, while a loss function is minimized, are you sure is a metric that minimizes? be specific, here C is what seems to be minimized, not yloss? or which model uses each loss if both are loss functions? what is really C and who minimizes/maximized? replace optimization by minimization/maximization. Also link original repo of Dice, Shap and then later your notebooks with use cases as well, everything in footnote urls.}

\subsubsection{DiCE use case: Predicting Bipolar Disease} 
%\nat{make sure thorough you use credit risk scoring and not credit scoring, may be a different thing}
We will focus on the use case presented in subsection \ref{use case shap}. As mentioned before, Bipolar disorder is a mental health condition that causes extreme mood swings that include emotional highs and lows. With the intent of supporting psychiatrists' decision making, we implement a ML model to determine whether a person has or not this disease. In order to create a system in which experts can trust, we need to provide explanations about how the model  determined a conclusion. 

In this particular use case, there is a particular interest in providing psychiatrists with the criteria that would change the model output (i.e., a change in the diagnosis of bipolar disease)\footnote{DiCE guide available at \url{https://colab.research.google.com/drive/12jw91RouPBc9slFwB2OWRwYB6Ckviiv3?usp=sharing}. We implemented DiCE using the framework developed by Mothilal et al. \url{https://github.com/interpretml/DiCE}}.

%%%%%%%%Corrected
In Table \ref{tab:dice examples}, we generate 2 counterfactual explanations for a given patient. The trained model (i.e., Random Forest) predicts a given patient has no bipolar disease. CF examples show which features should change in order to be diagnosed with bipolar disease. The example in Table \ref{tab:dice examples} shows that if the patient was older, the diagnosis (CF instance 2 in Table \ref{tab:dice examples}) would have been positive (i.e., Bipolar Disease).

%\nat{is this a word in the dictionary?? explain what it means in this context}
 \begin{table}[htbp!]
     \centering
     \footnotesize
     \begin{tabularx}{\linewidth}{c|c|c|c|c|c}
     \hline 
          Patient Data & Age & MADRS 1 & MADRS 2 & \makecell{Unipolar Depressive \\ Probability} 
          %\nat{healthy? see original paper I cite in first table to find proper term} 
          & \makecell{Bipolar II \\ Probability}\\ \hline \hline
          \textbf{Original Data} & 45-49   & 24 & 25 & 0.72 & 0.28 \\ \hline 
          CF instance 1 & \textbf{50-54} & 24 & \textbf{21.2} & 0.36 & 0.64 \\ \hline
          CF instance 2 & \textbf{65-69} & 24 & \textbf{20.5} &0.48 & 0.52\\ \hline
     \end{tabularx}
     \caption{Comparison between the features that DiCE modifies in order to change the model outcome from 0 (No Bipolar Disease) to 1 (Bipolar Disease) for a single patient (i.e. case). All other features remained fixed. % AYOUB: ToDo: \cnat{add accronmy complete from madrs 1 and 2 and what they are}
     }
     
     %\cnat{add outcome of model and probability of having BD (bipolar) to the table. What does marriage have to do here? show the full datapoint with all the features or at least the top k (fitting in table) most relevant. Is really marriage one of them?}.}
     \label{tab:dice examples}
 \end{table}

%\begin{figure}[hbtp!]
%    \centering
%    \includegraphics[width=0.4\textwidth]{figures/CF_explanations.png} 
%    \caption{Counterfactual (CF) examples generated using DiCE \cite{Mothilal_2020}. Green points are cases where the loan applicant would be approved. We selected a set of variables that are the easiest for the customer to vary (i.e., employment status, the behaviour on the payment of previous loans and the loan amount demanded). We generate 3 different CF examples for the consumer, but the amount of CFs to be generated can be decided by the user of DiCE or the applicant --if unsatisfied with previous CFs. However, the number of CF that can be generated is limited by the restrictions imposed to create new CF examples. % . The size of the CF depends mostly on the consumer.
%    }
%  %\nat{i.e., the most similar examples to the original input data that minimally changing its feature vector, change the model outcome. Ayoub indicate also how the amount of CFs are selected (is it arbitrary? say it here) and how many the user can ask for in the text and how these two axes are chosen wrt to all other possible features that could be used as axes. I.e., answer here in caption too: How did you query the system DiCE to choose the feaetures in x and y axes. }.}
%    \label{fig:CF examples}
%\end{figure}

\subsubsection{DiCE analysis: pros and cons}
Advantages of using DiCE include the agnosticism of the method, as a capability of generating high number of unique counterfactual explanations for any given ML model. It equally allows to produce explanations that are easily conveyable not only to developers but also to non technical audiences. On the other hand, currently, the disadvantage of DiCE is that works only for differentiable models, since it uses gradient  descent for the optimization process.
%assumes the knowledge of the gradient of the machine learning model.  This assumptions prevent the use of DiCE for full black-box models\nat{I dont understand, why are we showing it here in the guide then? DNNs are fully black box models}\footnote{DiCE original source: \url{https://github.com/interpretml/DiCE}. Guided DiCE guide and credit risk scoring example guide: \url{https://colab.research.google.com/drive/1nUTTTfcCuxsnZmaJpfvLsxRB4FFaORVK?usp=sharing}}.

\section{XAI techniques for image models} \label{images}

Convolutional Neural Networks (CNNs) constitute the state-of-art models in all fundamental computer vision tasks (image classification, object detection, instance segmentation) as of 2021. They are built as a sequence of convolutional and pooling layers that automatically learn and entails extremely complex internal relations between features. At the end of the sequence, one or multiple fully connected layers are used to match the output features map into scores.

While some XAI techniques try to delve inside the network and interpret how the intermediate layers see the external world, this guide presents a technique that try to understand the decision process of a Convolutional Neural Network (CNN) by mapping back the model output into the input space to see which parts of the image were discriminative for the prediction. This choice is motivated by the simplicity offered by visual understanding of explanatory elements that can be relevant for a wide audience.

\subsection{Grad-CAM use case: image classification}

Gradient-weighted Class Activation Mapping (Grad-CAM) \cite{selvaraju2016grad} uses the gradients (of any target concept) flowing into the final convolutional layer to produce a coarse localization map, highlighting the important regions in the image for predicting the concept. In other words, the Grad-CAM technique makes it possible to know which part of the image contributed the most to the model's prediction. 

In order to obtain the class-discriminative localization map $L^c_{Grad-CAM}$ for any class $c$, Grad-CAM computes the gradient $y^c$ of the score for class $c$ with respect to the feature map activation $A^k$ for feature map $k$ of a convolutional layer. These gradients are global-average-pooled by summing feature map activations $A^k_{i,j}$ over the width $i$ and height $j$ of the activation map containing $Z$ pixels to obtain the neuron importance $\alpha^c_k$, defined as:

\begin{equation}
    \alpha^c_k = \frac{1}{Z}\sum_i\sum_j \frac{\partial{y^c}}{\partial{A^k_{i,j}}}
\end{equation}

The output of Grad-CAM, heatmap $L^c_{Grad-CAM}$, is obtained by performing a weighted combination of forward activation maps with a Rectifier Linear Unit (ReLU) activation function. We usually normalize the heatmap and color it to make it more visually interpretable\footnote{Grad-CAM guide available at: \url{https://colab.research.google.com/drive/1ZXznvG_G1Y-JyHX9a_x6yKrXHhMp6tpm} adapted to a medical use case from \url{https://pyimagesearch.com/2020/03/09/} and \url{https://colab.research.google.com/drive/1bA2Fg8TFbI5YyZyX3zyrPcT3TuxCLHEC?usp=sharing} adapted to an additional domain from \url{https://keras.io/examples/vision/grad_cam/}}.

%removed for space purpose
\begin{equation}
    L^c_{Grad-CAM} = ReLU(\sum_k \alpha^c_k A^k)
\end{equation}

\begin{algorithm}
\caption{Grad-CAM Algorithm: Computing a class activation map as output explanation for a given classified image}
\begin{algorithmic}[1]
\REQUIRE Input Image $I$, Classifier $C$
\STATE Step 1: Isolate the last convolutional layer of model $C$
\STATE $LastLayer \leftarrow C.LastConvolutionalLayer$
\STATE Step 2: Create a model mapping the input image to the activations of the last layer
\STATE $ActivationMap \leftarrow LastConvModel(I, LastLayer)$
\STATE Step 3: Create a model mapping the activations of the last layer to the class predictions 
\STATE $PredMap \leftarrow PredModel$ $(LastLayer.Output, C.Output)$
\STATE Step 4: Compute activations of the last layer
\STATE $Activations \leftarrow ActivationMap(I)$
\STATE Step 5: Compute class predictions
\STATE $Predictions \leftarrow PredMap(Activations)$
\STATE Step 6: Compute the gradient of the top prediction
\STATE $TopPrediction \leftarrow Max(Predictions)$
\STATE $GradTopPrediction \leftarrow TopPrediction.Gradient$
\STATE Step 7: Multiply each channel in the activation map by the mean of the 
across the dimensions.
\STATE $PooledGradients \leftarrow Pool(GradTopPrediction)$
\FOR{$i \in Range(Activations)$}
\STATE $Activations[i] \leftarrow Activations[i] * PooledGradients[i]$
\ENDFOR
\STATE Step 8: Return the heatmap for class $C$ activation as the mean of the activation map:
\RETURN $ClassActivationMap \leftarrow Mean(Activations)$ 
\end{algorithmic}
\end{algorithm}

\begin{figure}[htbp!]
\centering
\includegraphics[scale=0.25]{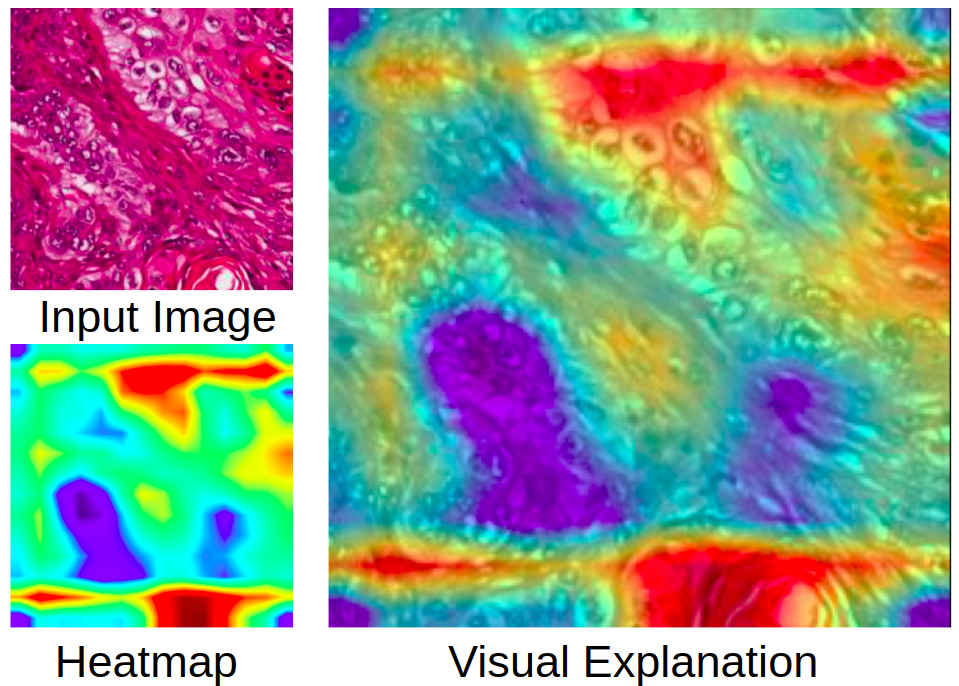} 
\caption{Grad-CAM example on a medical image representing an esophagus membrane and classified as such by a binary classification model using VGG16 trained on TCGA (Cancer Genome Atlas)\protect\footnotemark. Superimposed visualization of Grad-CAM's heatmap and the input image, showing the model mostly used the bottom and the top of the image in order to make its prediction.} %A Colab guide showing Grad-CAM functioning is available in \url{https://colab.research.google.com/drive/1bA2Fg8TFbI5YyZyX3zyrPcT3TuxCLHEC?usp=sharing}}
%\cnat{ Can you provide a ground truth segmentatino for this image to give context and provide any degree of evidence on how well it works? or find an image whwer this can be shown by gradcam vs segmentation GT image?}. Anna: As far as your question: Ground-truth segmentation :-) That would mean we need some doctors to do it. Of course, a segmentation done by software could be a good approximation of that. But this is an excellent use-case for the Kandinsky Patterns! The ground truth is built-in because we generated the synthetic data. So if we are to compare the Grad-CAM results and ground truth - that is our exploration playground to do so.
\label{fig:grad_cam}
\end{figure}

The goal of visualizing class activation maps in a CNN is to ensure that the model is taking a decision for the right reason and that it does not contain any inner bias due to learned spurious correlations or purposely misleading selected data. 
Taking as example Fig.\ref{fig:grad_cam}, a binary classifier was trained to classify pictures of Lymphoid tissue and Esophagus membrane. We want the model to be able to predict that the RGB input image represents an Esophagus membrane because it contains features typical of an Esophagus. An expert in the medical field would be able to verify that the hottest regions are the ones that should be used to make the prediction.

%and not because there is a grated carrot and a kitchen-like background since we want to be able to detect a rabbit even when there is no carrot nor any other typical background.
%This XAI technique may be suitable to verify the reliability of an animal classifier and help show if the model pays attention to the actual animal rather than to the background.

%\begin{figure}[htbp!]
%\centering
%\includegraphics[scale=0.5]{figures/rabbit_gradcam.PNG} 
%\caption{Superimposed visualization of Grad-CAM's heatmap and the input image, showing the model mostly used the center-right of the image (where the head of the rabbit is) in order to make its prediction.} %A Colab guide showing Grad-CAM functioning is available in \url{https://colab.research.google.com/drive/1bA2Fg8TFbI5YyZyX3zyrPcT3TuxCLHEC?usp=sharing}}
%\label{fig:grad_cam}
%\end{figure}

%

\footnotetext{available at the GDC Data Portal \url{https://portal.gdc.cancer.gov/}}

\subsection{Grad-CAM suitability analysis: pros and cons}

The Grad-CAM technique has the advantage of being easy to understand, as the explanation is visual, and easy to implement for any gradient-based model, as it does not require to modify the model architecture. However, the interpretation of the heatmap is subjective and therefore induces a human bias since the explanation does not come directly from the model but from an interpretation that the user makes, based on what it is able to recognize on the heatmap \cite{WhatDoesExplainableAImean}.
Also the Grad-CAM is class-discriminative but lacks the ability to show fine-grained importance as the heatmap is coarse and not in high-resolution. This means that it can determine globally which region contributed the most to detecting a certain class, but not precisely which pixels. Guided Grad-CAM \cite{selvaraju2017grad} proposes a high-resolution class-discriminative visualization by combining Grad-CAM with existing fine-grained visualizations. Nevertheless, the validity of explanations obtained by saliency-based techniques can be misleading \cite{adebayo2018sanity} as it was shown the relationship between \textit{good} saliency and generalization performance is tenuous that improved generalization is not always accompanied by improved heatmaps \cite{viviano2019saliency}.

\section{XAI Techniques for language models}  \label{text}

%In this section we investigate techniques to explain textual data. 
Most of the information available worldwide is in text form, from legal documents to medical reports. A variety of deep learning models have been applied to improve and automate complex language tasks. Examples of such tasks include, but are not limited to, tokenization, text classification, speech recognition, machine translation, and document summarizing.

%Due to the complexity of these NLP models, a strand of explainability research focuses on facilitating the identification of different features that contribute to their outputs \cite{kokhlikyan2020captum}. Natural Language Processing models' attempt to extract information and insights from natural language data. Danilevsky et al. \cite{danilevsky2020} clustered the XAI literature for NLP with respect to the type of explanation (i.e., post-hoc, self-explaining), and whether the information or justification for the model’s prediction concerns a specific input (i.e., local), or the functioning of the model as a whole (i.e., global). Their taxonomy identified five main explanations techniques, namely: feature importance \cite{voskarides2015}, surrogate model \cite{ribeiro2016modelagnostic}, example-driven \cite{croce2019}, provenance-based \cite{zadeh2018}, and declarative induction \cite{prllochs2019}. 

Among the existing Natural Language Processing (NLP) models, we analyzed transformers models for two pivotal reasons: they rely on the attention mechanism (i.e., initially designed for neural machine translation), and they are exceptionally effective for common natural language understanding (NLU) and natural language generation (NLG) tasks \cite{vaswani2017attention}. Transformer models are general-purpose architectures such as

The Transformer architecture works by weighing the influence of different parts of the input data, and aims at reducing sequential computation by relying entirely on the self-attention mechanism to compute a representation of its inputs and outputs \cite{vaswani2017attention}. 
In practice, the encoder represents the input as a set of key-value pairs, ($\mathcal{K}, \mathcal{V}$), of dimension $dk$ and $dv$ respectively. The decoder packs the previous output into a query $\mathcal{Q}$ of dimension $m$ and obtains next output by mapping this query against the set of keys and values \cite{weng2018attention}. The matrix of outputs, also called score matrix, determines the importance of a specific word with respect to other words.
The score matrix is the result of a %scale-dot
scaled dot-product where the weight assigned to each output is determined by the dot-product of the query and all keys (Eq. \ref{attention_transformer}).

%\nat{can we have also encoder equation if this is the decoder eq?}
\begin{equation}
\operatorname{Attention}(\mathcal{Q}, \mathcal{K}, \mathcal{V})=\operatorname{softmax}\left(\frac{\mathcal{Q K}^{\top}}{\sqrt{dk}}\right) \mathcal{V}
\label{attention_transformer}
\end{equation}

The attention mechanism repeats $h$ times with different, learned linear projections of the queries, keys and values to $dk$, $dk$ and $dv$ dimensions, respectively. The independent attention outputs of each learned projection are then concatenated and linearly transformed into the expected dimension \cite{weng2018attention}. 

\begin{equation}
\begin{aligned} \operatorname{MultiHead}(\mathcal{Q}, \mathcal{K}, \mathcal{V}) &=\text { Concat }\left(\operatorname{head}_{1}, \ldots, \text { head }_{\mathrm{h}}\right) \mathcal{W}^{O} \\ \text {where head}_{\mathrm{i}} &=\operatorname{Attention}\left(\mathcal{Q} \mathcal{W}_{i}^{Q}, \mathcal{K}\mathcal{W}_{i}^{K}, \mathcal{V} \mathcal{W}_{i}^{V}\right) \end{aligned}
\label{multi_head}
\end{equation}
In the multi-head attention (Eq. \ref{multi_head}), $h$ corresponds to the parallel attention layers (i.e., heads) and the $W$s are all learnable parameter matrices (i.e., $W_{i}^{Q} \in \mathbb{R}^{d_{\text {model }} \times d_{k}}$, $W_{i}^{K} \in \mathbb{R}^{d_{\text {model }} \times d_{k}}$, $W_{i}^{V} \in \mathbb{R}^{d_{\text {model }} \times d_{v}}$ and
and $W^{O} \in \mathbb{R}^{h d_{v} \times d_{\text {model }}}$). 

\subsection{An Example of Transformer Architecture: Bidirectional Encoder Representations from Transformers (BERT)}
\begin{figure}[htbp!]
\centering
\includegraphics[scale=0.13]{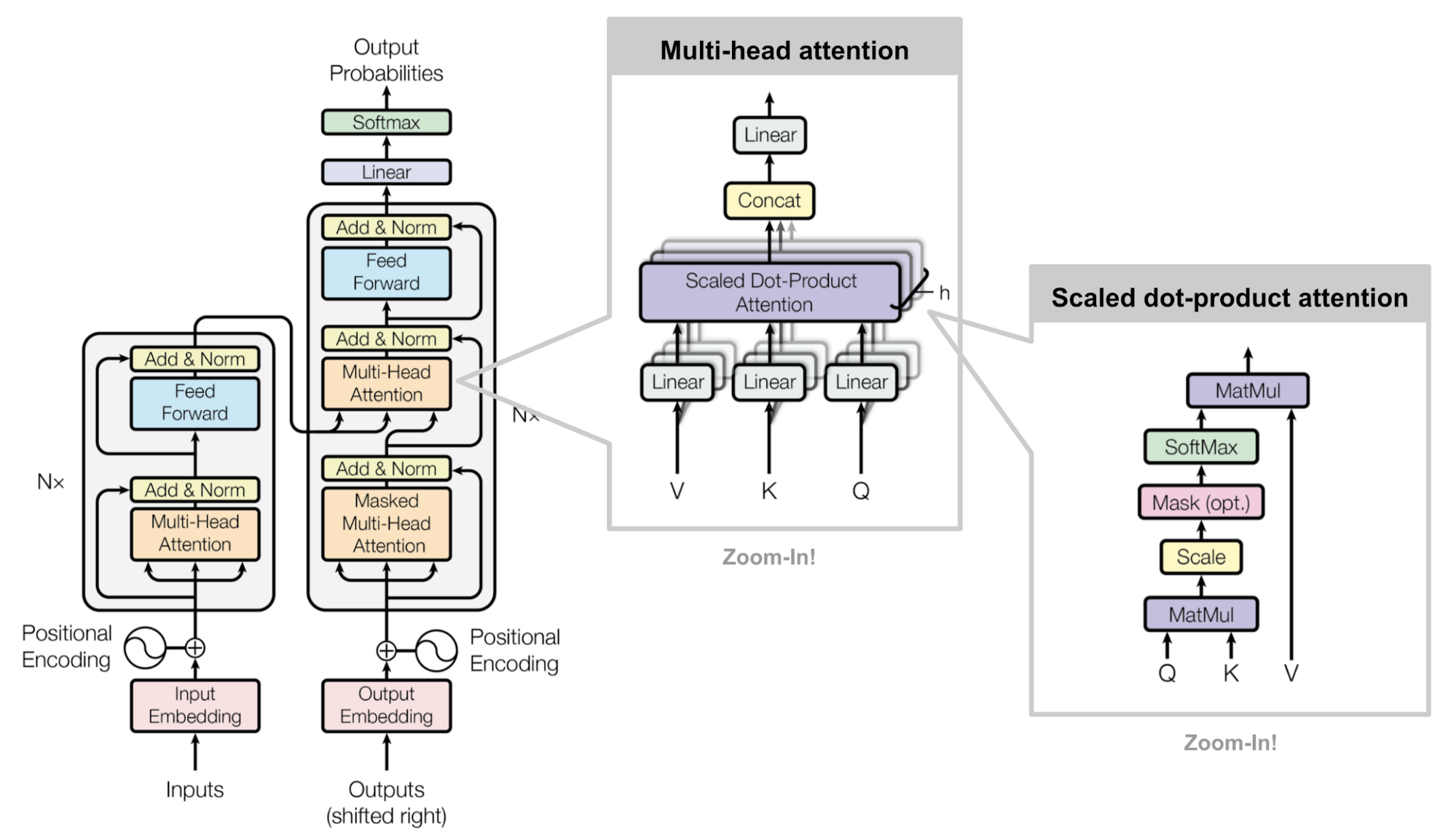}%todo remake this image on illustrator
\caption{High-level overview of the BERT Transformer model: the input is a sequence of tokens embedded into vectors, the output is a sequence of vectors linked by the index to the input tokens. The encoder multi-head attention mechanism computes queries, keys and values from the encoder states. The encoder feed-forward network takes additional information from other tokens and integrates them in the model. The decoder masked multi-head attention mechanism computes queries, keys and values from the decoder states. The decoder multi-head attention mechanism looks at the source of the target tokens taking the queries from the decoder states and the keys and values from the encoder states. The decoder feed-forward network takes additional information from other tokens and integrates them in the model (original image from \cite{vaswani2017attention}).}
\label{fig:BERT_overview}
\end{figure}

Bidirectional Encoder Representations from Transformers (BERT) \cite{devlin2019} is a transformer-based machine learning technique for NLP pre-training developed by Google. The peculiarity of this technique is that applies bidirectional training to language modeling.
%BERT Architectures of used for several NLP tasks. Examples of BERT downstream tasks are: sentence pair classification (e.g., MNLI, QQP), single sentence classification (SST-2, CoLA), question answering (e.g., SQuAD), single sentence tagging tasks (e.g., CoNLL-2003).
In contrast to directional models that read the text input sequentially (e.g., OpenAI GPT \cite{radford2019}, ELMo \cite{peters2018}), a bidirectional encoder processes the entire sequence of words at once. This way of processing words allows BERT to learn to unambiguously contextualize words based on both left and right words, and by repeating this process multiple times (i.e., multi-head), to learn different contexts between different pairs of words. Fig. \ref{fig:BERT_overview} describes the BERT Architecture. To define the goal of the prediction, BERT makes use of two techniques: Masked Language Modeling (MLM), inspired by the \textit{Cloze Procedure} \cite{taylor1953}, and Next Sequence Prediction (NSP) \cite{kiros2015}. The former consists of substituting approximately the 15\% of the tokens with a mask token and querying the model to predict the values of the masked tokens based on the surrounding words. The latter involves training the model by giving pairs of sentences as input to learn to predict whether the second sentence in the pair is the subsequent sentence in the original document.

Due to the increased attention received by the Transformer models, there exist a number of interfaces for exploring the inner workings of transformer models.
\textit{Captum}\footnote{Captum: \url{https://github.com/pytorch/captum}} is a multi modal package for model interpretability built on PyTorch. \textit{Captum} attribution algorithms can be grouped in three main categories; primary, layer and neuron attribution algorithms. Primary attribution algorithms allow us to attribute output predictions to model inputs. Layer attribution algorithms allow us to attribute output predictions to all neurons in the hidden layer. Neuron attribution algorithms allow us to attribute an internal, hidden neuron to the input of the model \cite{kokhlikyan2020captum}.

%\nat{language model? withe fully once, first time introduced} 

\subsection{Explaining Transformer Models with Transformer Interpret.}
%%%

%Together with other libraries (e.g., Ecco, Flair, Interpret-Flair\footnote{Ecco: \url{https://github.com/jalammar/ecco}, Interpret-Flair: \url{https://github.com/robinvanschaik/interpret-flair}, FlairNLP: \url{https://github.com/flairNLP/flair}.}), \textit{Captum} has been broadly used to explain Transformer models. Inspired by Captum and Hugging Face, 

%Transformers Interpret available here: \url{https://colab.research.google.com/drive/12_Q1WI05oXMfG-B_GwyiLjyU-rxRyFdX?usp=sharing} is a dedicated tool for interpreting Transformer Models. 
The Transformer-Interpret tool\footnote{\url{https://colab.research.google.com/drive/12_Q1WI05oXMfG-B_GwyiLjyU-rxRyFdX?usp=sharing}} has the advantage of relying on two well documented packages and frameworks (e.g., Captum and HuggingFace Transformers). It provides simple methods to explain most common natural language processing tasks performed by Transformer models, such as sequence classification, zero-shot classification, and question answering. The default attribution method used by Transformer-Interpret is Integrated Gradients (IG) \cite{sundararajan2017axiomatic}. Integrated Gradients visualize the importance of the input feature in the model's prediction. To do so, IG computes the integral of gradients with respect to inputs along the path from a given baseline to input. %\ref{integratedgradients}.

The integrated gradient along the $i^{t h}$ dimension for an input $x$ and baseline $x\prime$ is defined as follows.

\begin{equation}\label{integratedgradients}
    IntegratedGrads _{i}(x):=\left(x_{i}-x_{i}^{\prime}\right) \times \int_{\alpha=0}^{1} \frac{\partial F\left(x^{\prime}+\alpha \times\left(x-x^{\prime}\right)\right)}{\partial x_{i}} d \alpha
\end{equation}
Where $\frac{\partial F(x)}{\partial x_{i}}$ is the gradient $F(x)$ along $i^{t h}$ dimension. $\alpha$ is the scaling coefficient. 
%The equations are copied from the original paper \cite{sundararajan2017axiomatic}.

IG's aims to satisfy two desirable axioms for an attribution mechanism:
\begin{itemize}
    \item \textit{Sensitivity.} If a modification in a feature's value leads to a change in the classification output, then that feature should have a non-zero attribution as it means this feature must have played a role in the classification. 
    \item \textit{Implementation Invariance.} The attribution method result should not depend on the parameters of the neural network, i.e. two neural networks giving the same output for a certain input should have the same attribution even if their weights are different.
\end{itemize}

%why is it called integrated,
%maybe add an algorithm and an example on how to compute integrated gradients based on some input features

Early interpretability methods for neural networks assigned feature importance scores using gradients because computing the gradients of the input with regard to the output is \textit{Implementation Invariant} but does not satisfy \textit{Sensitivity} as a feature change does not necessarily yield a non-zero gradient for that feature. IG adds \textit{Sensitivity} to gradients based methods by establishing a baseline (a reference input with an equiprobable prediction between the different classes) and compute the sum of all the gradients from this baseline to the input of interest (the one we try to explain). This allows to know how the change of value of a feature of the input leads to a change in the classification.

\begin{figure}[htbp!]
\centering
\includegraphics[scale=0.20]{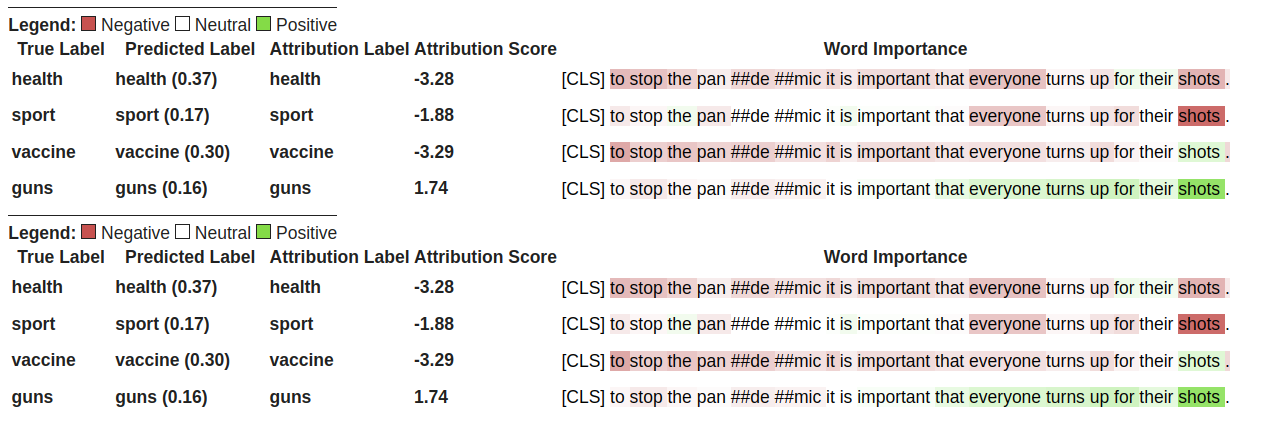}
\caption{Zero-Shot Classifier on COVID-19 Content from Twitter. Data from \cite{muller2020covid}.}
\label{fig:sentiment}
\end{figure}

A commonly text classification task is sentiment analysis. Sentiment analysis has the objective of detecting positive, neutral or negative sentiment in text. Transformer-Interpret uses IG to identify and visualize how positive/negative attribution numbers associated to a word contributes positively/negatively towards the predicted class, as shown in Fig \ref{fig:sentiment}.

\section{XAI techniques to explain image, text and graph classification models: Layer-wise relevance propagation (LRP)} \label{LRP}

Layer-wise Relevance Propagation (LRP) is a method that produces a heatmap for every input data sample \cite{montavon2017explaining}. The heatmap's data structure and size is the same as the input's and its highlighted parts denote the areas of the input that played the highest role in the classification. LRP is an XAI method that applies to several neural network architectures and thereby to several types of data that they can process. Each subsection next deals with a type of data, showing that LRP is flexible enough to handle different data modalities.

LRP's methodology is based on the Taylor expansion of a function $f(x)$ at point $a$, as expressed by equation \ref{Taylor_1}:

\begin{equation}
f(x) = f(a) + \frac{f'(a)}{1!}(x-a) + \frac{f''(a)}{2!}(x-a)^2 + \cdots
\label{Taylor_1}
\end{equation}

Provided that a neural network is computing a non-linear function $f(\mathbf{x})$ of its input $\mathbf{x}$, the function can be expanded near a root point $\tilde{\mathbf{x}}$. The higher order terms can be considered negligible and represented by a constant $\epsilon$.

\begin{equation}
f(\mathbf{x}) = f(\tilde{\mathbf{x}}) + \Bigg( \frac{\partial f}{\partial \mathbf{x}} \biggr\rvert_{\mathbf{x} = \tilde{\mathbf{x}}} \Bigg)^T
  (\mathbf{x} - \tilde{\mathbf{x}}) + \epsilon = 0 + \sum_p \underbrace{\frac{\partial f}{x_p} \biggr\rvert_{\mathbf{x} = \tilde{\mathbf{x}}} (x_p - \tilde{x}_p)}_\text{$R_p(\mathbf{x})$} +  \epsilon 
  \label{Taylor_2}
\end{equation}

Since $f(\tilde{\mathbf{x}}) = 0$, and assuming without loss of generality that $\tilde{\mathbf{x}}$ is an image composed by pixels $p$, one can re-write Eq. \ref{Taylor_2} as follows:

\begin{equation}
f(\mathbf{x}) = 0 + \sum_p \underbrace{\frac{\partial f}{x_p} \biggr\rvert_{\mathbf{x} = \tilde{\mathbf{x}}} (x_p - \tilde{x}_p)}_\text{$R_p(\mathbf{x})$} + \epsilon 
  \label{Taylor_3}
\end{equation}

The goal of LRP is to redistribute the neural network output onto the input variables; i.e., the relevance $R_j$ to lower-level relevances $\{R_i\}$. Starting from the output layer, one can restate Eq. \ref{Taylor_2}:

\begin{equation}
\begin{array}{l}
\sum_j R_j = \Bigg( \frac{\partial (\sum_j R_j)}{\partial{\{x_i\}}}\biggr\rvert_{\partial \{\tilde{x}_i\} } \Bigg)^T (\{x_i\} - \{\tilde{x}_i\}) + \epsilon = \\ 
\sum_i \sum_j \frac{\partial R_j}{\partial x_i}\biggr\rvert_{\partial \{\tilde{x}_i\} } (x_i - \tilde{x}_i) + \epsilon 
\end{array}
\label{Taylor_4}
\end{equation}

\begin{align}
\sum_j R_j &= \Bigg( \frac{\partial (\sum_j R_j)}{\partial{\{x_i\}}}\biggr\rvert_{\partial \{\tilde{x}_i\} } \Bigg)^T (\{x_i\} - \{\tilde{x}_i\}) + \epsilon \\ 
&= \sum_i \sum_j \frac{\partial R_j}{\partial x_i}\biggr\rvert_{\partial \{\tilde{x}_i\} } (x_i - \tilde{x}_i) + \epsilon 
\label{Taylor_5}
\end{align}

One of the challenges of LRP is to find a neighbouring point $\tilde{\mathbf{x}}$ of $\mathbf{x}$, for which $f(\tilde{\mathbf{x}}) = 0$ (root point). A good root point is one that removes the elements of a datapoint $\mathbf{x}$ that cause $f(\mathbf{x})$ to be positive. For example, in the case of object detection and a classifier that discriminates between images containing an object and images that do not, an optimization method should look for a similar image that contains an object not recognizable from the classifier - hence the output $f(\tilde{\mathbf{x}}) = 0$. Examples of such images are some that contain blur or have parts that are relevant for the recognition of the object replaced with gray/black (non-informative) pixels.

%ToDo: Is Grad-CAM applicable only to CNNs? Not exactly, as shown here: https://openaccess.thecvf.com/content_CVPR_2019/papers/Pope_Explainability_Methods_for_Graph_Convolutional_Neural_Networks_CVPR_2019_paper.pdf, Graph Convolutional Neural Networks can also be profited from that method. Added below:

The main difference between Grad-CAM and LRP is that although both of them compute gradients, the first one computes the gradient concerning the feature maps activations of CNNs, whereas the second one does it also for other types of architectures, not necessarily CNNs, and it is done in a per-neuron basis. Although also a heatmapping method, Grad-CAM does not compute relevances per se, but rather tries to locate the part of the input image that is responsible for the predicted label. Grad-CAM is only applied onto that architecture (with an exception of Graph Convolutional Neural Networks \cite{pope2019explainability}), whereas LRP has shown to be more universal. 

\begin{figure}[htbp!]
\centering
\includegraphics[width=8cm,height=5cm]{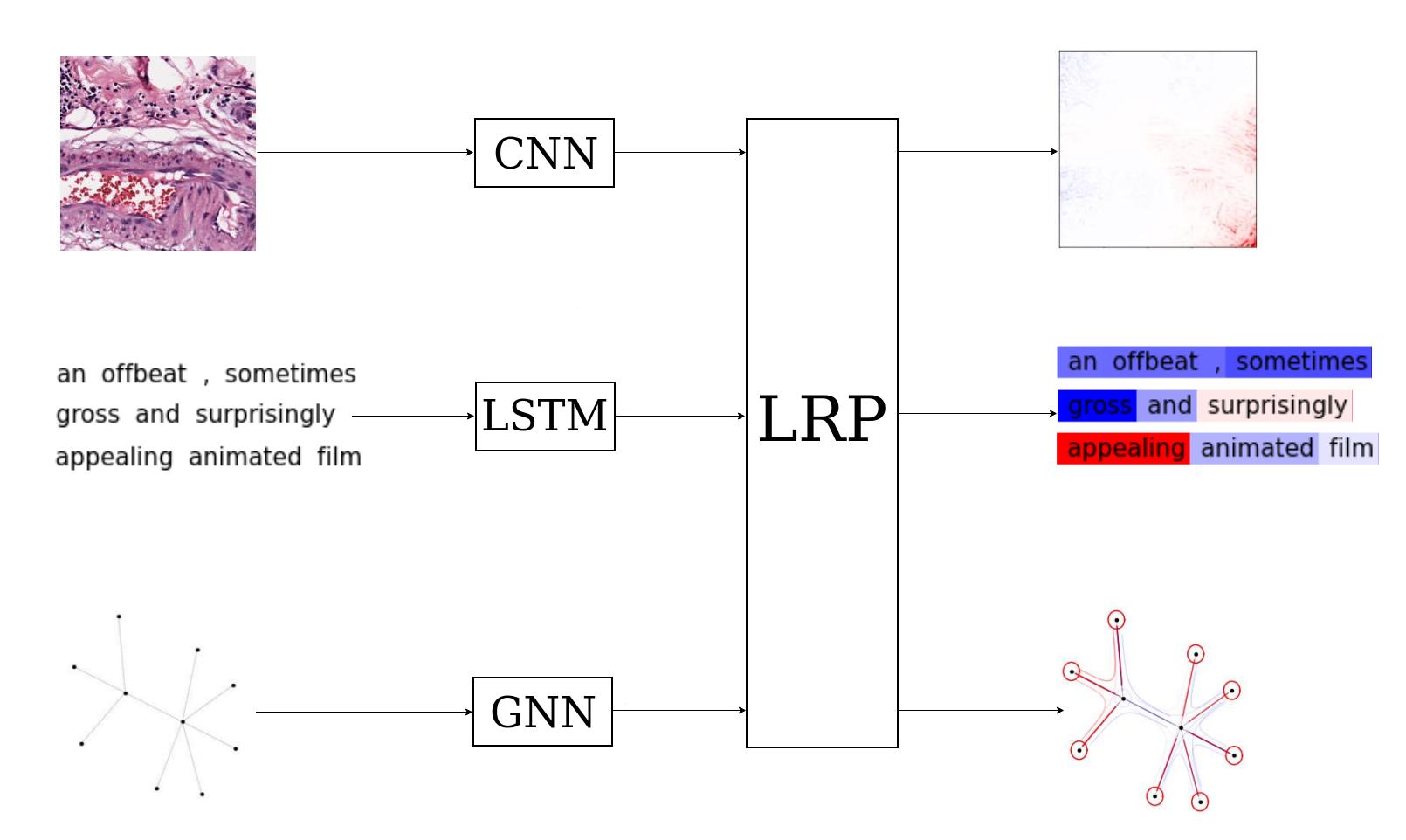}
\caption{Layer-wise Relevance Propagation (LRP) applied onto various neural network architectures, each of them processing different types of data.}
\label{fig:LRP_Multimodal}
\end{figure}

% % % % % % % % % % % % % % % % % % % % % % % % % % % % % % % % % % % % % % % % % %
\subsection{LRP applied onto Fully Connected (FC) Neural Network that solves a regression problem}

In this task\footnote{Notebook on LRP for a Fully Connected (FC) Neural Network using synthetic data: \url{https://colab.research.google.com/drive/1Md2Rz3Ff1r05zq98cYndiEqrhg-DGYPv?usp=sharing}}, the interest is in computing the
relevance at particular elements of a small, fully connected neural network. The network consists of only one input layer (its neurons are indexed by ($i$), one hidden ($j$) and one output layer ($k$). $x_i$ represents the values of the input neurons, $x_j$ the outputs of the hidden layer neurons and $x_k$ the outputs of the output layer neurons. The nonlinear function of the hidden layer is the ReLU, which is expressed by the equation $x_j = max(0, \sum_i x_i w_{ij} + b_j)$. $w_{ij}$ are the weights between the $i$-th and $j$-th layer and $b_j$ the bias (omitted in this task). The nonlinear function performed by the output layer is the sum pooling function, expressed by $x_k = \sum_j x_j$. 

\begin{figure}[htbp!]
\centering
\includegraphics[width=8cm,height=4.75cm]{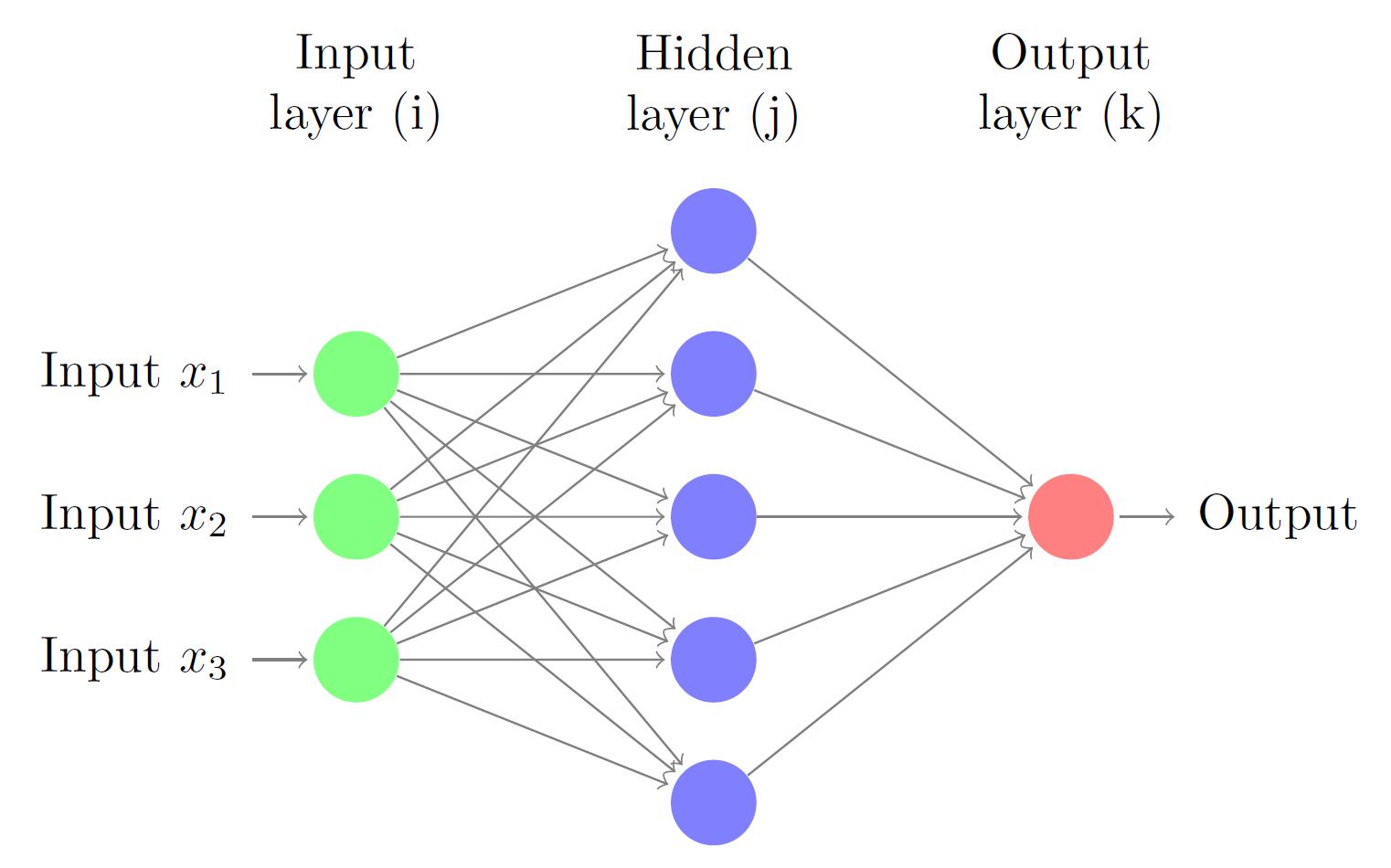}
\caption{Fully connected neural network that is used in the first task. It is composed by $3$ neurons in the input layer, $5$ neurons in the hidden layer and $1$ neuron in the output layer. The activation function that was used is the Rectified Linear Unit (ReLU)}
\label{fig:FC_Neural_Network_A}
\end{figure}

The fully connected neural network that is used in the first task is depicted in figure \ref{fig:FC_Neural_Network_A}. Relevances of each neuron at each layer (indexed by $i, j $ and $k$ correspondingly) are computed by the following set of equations:

\begin{align}
R_k &= x_k = \sum_j x_j, \\
\quad R_j &= x_j = max(0, \sum_i x_i w_{ij} + b_j)
\label{eq:lrp_output_hidden_input_layer}
\end{align}

\begin{equation}
\quad R_i = \sum_j \frac{w_{ij}^2}{\sum_{i'} w_{i'j}^2} R_j
\end{equation}

% \nat{if relevance is computed for the input or input layer? [specify when each one is applied as they all compute Rk, Rj and Ri, not clear each one when?] }
%\begin{equation}
%R_j = x_j = max(0, \sum_i x_i w_{ij} + b_j)
%\label{eq:lrp_hidden_layer}
%\end{equation}
% \nat{if j is the last layer (i.e., activation function?}
%\begin{equation}
%R_i = \sum_j \frac{w_{ij}^2}{\sum_{i'} w_{i'j}^2} R_j
%\label{eq:lrp_input_layer}
%\end{equation}
where $i'$ denotes all neurons of the input layer, including the $i$-th neuron. In this equation $i$ corresponds to one particular neuron in the input and $i'$ is an index over all of them. 

The relevance of the neuron in the output layer is completely specified by the sum of its inputs, since this is its functionality. The relevance of each neuron in layer $j$ is derived by using the equations \ref{Taylor_3} and \ref{Taylor_5}: $R_j = R_k(\tilde{\mathbf{x}}) + \frac{\partial R_k}{\partial x_j} \bigr\rvert_{\{ \tilde{x}_j \}} \cdot (x_j - \tilde{x}_j) = x_j = max(0, \sum_i x_i w_{ij} + b_j)$. Since the ReLU nonlinearity is used, it is ensured that $\{ \forall j: \tilde{x_j} \geq 0 \}$ and $\frac{\partial R_k}{\partial x_j} = \frac{\partial \sum_j x_j}{\partial x_j} = 1$. Therefore, the root point $\tilde{\mathbf{x}}$, for which $R_k(\tilde{\mathbf{x}}) = 0$ is $\tilde{\mathbf{x}} = {0}$. The computation of the relevances $R_i$ of each neuron in layer $i$ has a derivation that is out of scope for the purposes of this paper. Nevertheless, it is important to note that the relevance is proportional to the squared weight of the connection - having in mind that weights can take both negative and positive values.  

The relevance of the neurons of each layer (in general) is computed by using the relevances of neurons of the next one. 

% \nat{then what is the difference btw i and i'? is Ri always computed based on precedent layers to i or posteriors? indicate in each equation}

% \nat{formula 6 you. mean? be explicit. Also shouldnt they be called different not just Rk, Rj and Ri? hard to know when apply each} --- Each of them corresponds to the layer i (input), j (hidden), k (output). Now that we have space to add the figure, hope it's clearer.

% amplify \nat{what does it mean?}. It is explained by an example with positive input value and high positive weight.

% \nat{how is this found later, i.e. the highest contribution? it is not clear from the formulas that come next}. Since those inputs were amplified the most by positive weights (or negative inputs by large negative weights), they will have (comparable to the ones "suppresed" by values near zero) higher relevance. 

% random weights do not show any preference \nat{you mean 'do not display high activations'?} - unless they \nat{ their values?} are high (relative to others). 
% What I mean is that randomly distributed weights will not increase or suppress any input value in general. This state is the same as an untrained neural network, where the network has not learned anything. The prediction will be random, and the relevance heatmaps will not have any recognizable structure.

\begin{figure}[htbp!]
\centering
\includegraphics[width=9cm,height=4cm]{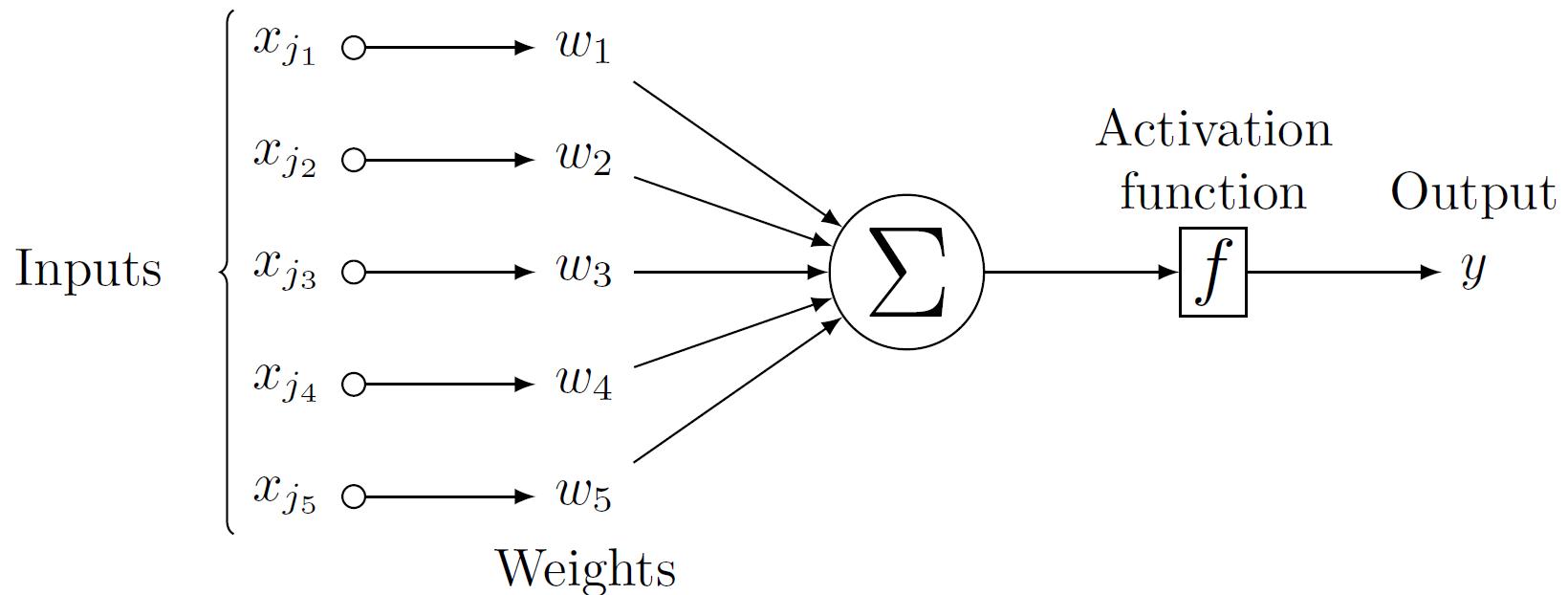}
\caption{The functionality of a neuron of a fully connected neural network with five inputs, five weights, the sum of their corresponding multiplication and the application of a nonlinear activation function.}
\label{fig:FC_Neural_Network_B}
\end{figure}

The overall goal of the task is to understand how the interplay between values of the input and the network weights define the computed relevance values. To better explain this, Figure \ref{fig:FC_Neural_Network_B} depicts the functionality of a neuron in a fully connected neural network. For convenience and without loss of generality we can think that this is the the neuron in the output layer $k$ of the task (see also \ref{fig:FC_Neural_Network_A}). Each input $x_{j1}$ to $x_{j5}$ will be multiplied by a corresponding weight ($w_1$ to $w_5$), and then the sum of all those multiplications $x_{j1} w_1 + \cdots + x_{j5} w_5$ comprises the input of the nonlinear activation function. In this case, the Rectified Linear Unit (ReLU) was used.  

For a fixed input, the exercises deal with two cases: the first one sets weights that are not randomly chosen. Thereby, positive weights multiplied by positive input values will generate a high positive value which in turn will consist the input to the activation function. Those parts will be, in retrospect, the ones that will have higher relevance in general. On the contrary, weights near zero will ``supress'' all highly positive or highly negative input values. The second case is quite the opposite of that; random weights do not show any preference to input values. By those means, one can compare the situation of a trained vs. not trained neural network that has randomly initialized weights.

The overall goal of the task is to understand how the interplay between values of the input and network weights defines the computed relevance values. When the weights are randomly distributed, only the highest input values will manage to create a high activation (with few exceptions). When the weights are not randomly distributed, then some relatively high values might be suppressed (by multiplication with a small weight) and not create a high activation. On the other hand,  some relatively small values, when multiplied with a high weight value, might induce a high activation. By those means, one can compare the situation of a trained vs. not trained neural network.

It is important to emphasize the difference between the backpropagation procedure that happens several times during the training of a neural network, and 
the backpropagation of relevance happens only once after the training is accomplished. Furthermore, two properties of LRP are used for verification of the computations with unit tests \cite{Saranti:2020:PropertyBasedTesting}, namely positivity and conservation (Eq. \ref{eq:positivity_and_negativity}) of relevance of the neurons at each layer:

\begin{equation}
\forall x, p: R_p(x) \geq 0, \quad \sum_i R_i = \sum_j R_j
\label{eq:positivity_and_negativity}
\end{equation}
%\begin{equation}
%\sum_i R_i = \sum_j R_j
%\label{eq:conservation}
%\end{equation}
where $x$ is the input, $p$ represents any neuron of the network, and layer $i$ precedes layer $j$.

\subsection{Explaining a GNN performing node classification on graphs with GNN-LRP}

Graph Neural Networks (GNNs) perform three main types of tasks on graph datasets: node classification, link prediction and graph classification. They can be thought of as an extension of Convolutional Neural Networks (CNNs) that processes non-grid structured data, therefore the filters cannot operate by the same means. %Lots of applications using biological, social and traffic data emerged in the last few years using GNNs because of the underlying structure of the data. 
% On the other hand, traditional problems of image and text processing can be handled by GNNs after a transformation of the corresponding data to graphs (e.g., image to scene graph).

%Since a graph is not an \textit{ordered} data structure (e.g. unlike images in which pixels have a precise order), the convolution function is expanded to a message-passing scheme. To learn the parameters of the network, typically one aggregates information from the neighbourhood of a node in an iterative manner. Afterwards, this information is combined and a non-linear function (which is shown to define the discrimination capabilities of the network) is applied. 

One of the simplest architectures is called GCN (Graph Convolutional Network) \cite{Schnake:2020:GNN-LRP}. The rules for aggregation and combination (Eq. \ref{GCN_aggregation_combination}) of the information lying in the features of the neighboring nodes and edges are: 

\begin{equation}
\bm{Z}_t = \bm{\Lambda} \bm{H}_{t-1}, \quad \bm{H}_t = \rho(\bm{Z}_t \bm{W}_t)
\label{GCN_aggregation_combination}
\end{equation}
where $t$ denotes the layer,%\footnote{$t$ here represents the neighborhood of the graph - sometimes referred to as the \textit{k-hop} neighborhood.}, 
and 
$\bm{\Lambda}$ is the Laplacian matrix of the input %\nat{input or computed? put an example of what kind of graph can be used here, e.g. a knowledge graph or protein interaction graph?} 
graph, which can be a scene, protein interaction, social media, a knowledge graph, etc. $\bm{\Lambda}{H}_{t-1}$ is the representation of the previous layer, $\bm{W}_t$ are the weights and $\rho$ is the non-linear activation function. The GNN-LRP method \cite{Schnake:2020:GNN-LRP} applies constraints (piecewise linear and positive homogeneity) to this nonlinearity (here ReLU is used). %For the LRP rules to be computed, 
Eq. \ref{GCN_aggregation_combination} is re-written and the GNN-LRP rule for computing the relevance $R_{jKL\dots}$ %\nat{complete: between [] j and [] L passing by K? what is J and K?}becomes:
% \nat{of neuron j for the edge connecting K and L becomes:} 
of neuron $j$ after one has processed nodes $K$ and $L$ by all neurons $k$ that gathered information from node $K$ becomes:

%R_kL: The relevance of neuron k after one has "visited"/processed node L ...
% \nat{of neuron x? or? formulate it as Rjkl. I dont undrestand why some indexes in eq 14 are capitalized and others not, if all represent neuron indexes? also shouldnt jkl preserve an order from j to l passing by k? }

\begin{equation}
R_{jKL\dots} = \sum_{k \in K}\frac{\lambda_{JK} h_{j} w_{jk}^\wedge}{\sum_J \sum_{j \in J} \lambda_{JK} h_{j} w_{jk}^\wedge } R_{kL\dots}
\end{equation}
where $K, L$ are elements of a walk on the input graph\footnote{A walk on the graph involves nodes that are processed by corresponding neurons at layers labeled %by the same means.
with the same (non-capital) indexes.} % \nat{walk between which nodes? this concept is not introduced yet, random walk? or mere neurons connecting weight?} - \anna{this is not a random walk, it is a walk on the graph so an element can be a node or edge that is processed by corresponding neurons at layers which are labeled correspondingly}
processed by neurons with $k, l$ indexes %\nat{
(capital letter subindexes represent nodes, while at the same time they also denote all neurons with that corresponding non capital index that process those nodes). Weight $w_{jk}^\wedge$ is a weighted sum (denoted by the $\wedge$) of the elements of matrix $\bm{W}_t$ that links neuron $j$ to neuron $k$ parameterized by a user-provided parameter $gamma$, which for different values, facilitates explanations with different visual details. $\lambda_{JK}$ is the element of the Laplacian matrix $\bm{\Lambda}$ corresponding to the connection between nodes $J$ and $K$, $h_{j}$ is the activation of neuron $j$, $R_{kL\dots}$ is the relevance of neuron $k$ after one has processed node $L$. The "$\dots$" indicate that the walk contains in general further nodes.
% \nat{and lambdaKL between layer? K and neuron? layer? L is...? mention and name every constant and what is each capital and non capital letter}

The task of graph classification contains three parts. At first, Barabasi-Albert graphs are created by the user. These are random scale-free networks that have a preferential attachment mechanism with user-defined growth factors. Nodes and edges can represent anything that can be approximated by a scale-free graph (a graph that has a degree distribution according to the power law). Examples of such graphs are citations' and social networks. This growth factor will be the label for the prediction. The GNN that will be used for this prediction is defined in the second part, along with the corresponding functions for training and computing relevances. The third part deals with training the network, use a test set to compute its performance and then applying GNN-LRP to compute and display a plot showing the \textit{walk's} relevances. 

The most important goal of this task is to understand what the computational path of a GNN is, and its recursive nature, to comprehend how this leads naturally to the relevance of walks in the graph. Our notebook\footnote{Notebook on LRP for an GNN trained on graph data \url{https://colab.research.google.com/drive/166FYIwxblfrEltkYqY_jiJoAm9VLMweJ?usp=sharing}} is a slightly changed version of %a notebook that is already provided from the
the researchers' original's\footnote{GNN-LRP: \url{https://git.tu-berlin.de/thomas_schnake/demo_gnn_lrp}}. The task is studying whether the % \nat{neuron? layer? class?}
walk relevances are plausible, and how they change if the length of the walk on the graph %\nat{between neurons connecting input and output?}
is changed, in juxtaposition with an adaptation of the GNN layer sizes (changing its amount, and size in term of neurons) with the aim of obtaining a model output and output explanation that are consistent with the ground truth and the expectations of a domain expert. 

For example, let us assume we have a scene graph of a medical image with cells and links. Cells are represented by the nodes and have various features such as size, color, or shape. It might be that a classification lies on the relevance of the size of a particular node of the graph. Paths that contain this node will have high relevance, but we will not know which node exactly it is and which feature exactly it is. GNN-LRP might not be the most adequate XAI method for this use-case.

One important challenge of all XAI methods is the explainability of misclassified samples. LRP does not at the moment provide a substantial and quantifiable benefit in comparison with other heatmapping methods; nevertheless, the perturbation analysis deals with misclassified samples in a more robust way than other methods because the performance is influenced (i.e., drops) monotonically after the removal of the relevant elements in sorted order. 

%This notebook is available online\footnote{\url{https://colab.research.google.com/drive/166FYIwxblfrEltkYqY_jiJoAm9VLMweJ}}.

% (\nat{because LRP-GNN could provide walk paths that are not? -what is done in this case? briefly mention}  -\anna{That's sooo subjective, sometimes the heatmaps or any type of explanation can be criticized from domain experts as being non plausible.}) 
% \nat{with what criteria would this be done?} - \anna{the students can change the number of GNN layers and add or remove neurons from them - I hope I understand the question correclty}. 
% \nat{I mean: specify when to stop changing configurations and based on what (qualitative/quantitative) measure?}
%I meant is there a more qualitative evaluation that observing configurations until having a more visually convincing map? Anna: Like whatkd of evaluation? Sorry, I don't get this :-(}   % \nat{and LRP helps in which way?} - \anna{it does not atm, particularly not in comparison with other heatmapping methods. It does not have a benefit over other methods in this regard; its only in the perturbation analysis (covered by the LSTM section) that its better. We can extend this task to do this too here; but I encourage students to see heatmaps of missclassified samples either way.}.

% % % % % % % % % % % % % % % % % % % % % % % % % % % % % % % % %
\subsection{Comparison of LRP with other XAI algorithms}

One of the major benefits of LRP is arguably the existence of both positive and negative relevance values \cite{Samek:2016:InterpretingPredictionsLRP}. The perturbation analysis looks enormously different with a method like Sensitivity Analysis (SA) \cite{Baehrens:2010:SensitivityAnalysis_1}. %\cite{Simonyan:2013:SensitivityAnalysis_2}. 
In practical terms, there is no differentiation of which parts of the input enhance the certainty of a prediction, which ones are an indication of another class (if we are dealing with a classification problem), and which parts are rather neutral (ideally the background). An explanation with SA labels both highly positive and highly negative relevant areas as important, without providing this differentiation which is of immense importance for interpretation.

As we are moving to the era of big data, it becomes necessary to deal with large amounts of data. Since most heatmapping XAI methods (including LRP) produce one heatmap per input sample, one can imagine that data scientists and domain experts do not have the time to go through that many heatmaps to understand and improve the behavior of the model. Therefore Semi-automated Spectral Relevance Analysis was invented \cite{Lapuschkin:2019:UnmaskingCleverHansPredictors}. This method was inspired from the detection of a so-called ``Clever-Hans'' effect in a high performing neural network that was classifying images of the PASCAL VOC2007 data set. The inventors of LRP found out that several images of the class ``horse'' contained a tag on the lower right side. This is an artefact that was also discovered in other datasets and is usually representative of the photographer and the camera. The neural network learned that this tag was so indicative for accurate prediction of the class ``horse'' that even experiments with images with cars and this tag were classified as ``horse''. To be able to semi-automatically separate between explanations that classify an image as ``horse'' because it actually contains a horse and those that contain some artefact, one can cluster the LRP heatmaps. If the resulting clusters are relatively far from each other, then one can assume that the neural network has found different ways to classify those images and maybe some of them are Clever-Hans. With human investigation of the heatmaps in each cluster, we can inspect the different strategies that the neural network has come up for accomplishing the task and take action as to improve both the model and the dataset. LRP is one of the first methods that opened the path for actionable insights and Actionable XAI (AXAI) \cite{Pfeifer:2021:NetworkModuleDetectionMultiModal}.

%%%%%%%%%%%%%%%%%%%%%%%%%%%%%%%%%%%%%%%%%%%%%%%%%%%%%%%%%%%%%%%%%

%---------------------------------------------------------------------------------
\section{Neural-symbolic AI for interactive explainability in Neural Networks} \label{Nesy}% \nat{in natural language?/to produce logic explanations? (try be a bit more concise on the purpose)}}

Neural-Symbolic Learning and Reasoning seeks to integrate principles from neural network learning with logical reasoning. Symbolic systems operate on a symbolic level where reasoning is performed over abstract, discrete entities, following logical rules. A common goal of %work on 
symbolic systems is to model %(certain aspects of) common-sense
reasoning, which inherently allows for better explainability. Neural networks, on the other hand, operate in the sub-symbolic (or \textit{connectionist}) level. Individual neurons do not necessarily represent a readily recognisable concept or any discrete concept. 

The integration between both levels can bridge low-level information processing --such as the one frequently encountered in perception and pattern recognition-- with reasoning and explanation on a higher, more cognitive level of abstraction. Realising this integration facilitates a range of benefits, such as achieving representations that are abstract, re-usable, and general-purpose. Having these readily available could concretely tackle some of the pressing issues with current deep learning practices.

\subsection{XAI in Neural-Symbolic AI}

Explainability in neuro-symbolic systems has been traditionally approached by learning a set of symbolic rules, known as Knowledge Extraction, and evaluating how well it may approximate the behaviour of a complex neural network by measuring the percentage of matching predictions on a test set, referred to as fidelity \cite{White2020}.

This is comparable to most contemporary explainability methods that are not powerful enough to guarantee the soundness and completeness of the explanation concerning the underlying model. Most metrics currently in place are lacking a reliable way of expressing this uncertainty. %For instance, \ivan{missing sentence}

The measured fidelity is supposed to be a good proxy of the closeness of the representation to the underlying model. However, this metric is limited in its capacity and ability to find semantically meaningful representations that allow for transparent reasoning, as it is solely optimizing for the resemblance of the explained model.

\begin{figure}[h!]
\centering
\includegraphics[trim={0 0 0 4cm},clip,width=\columnwidth]{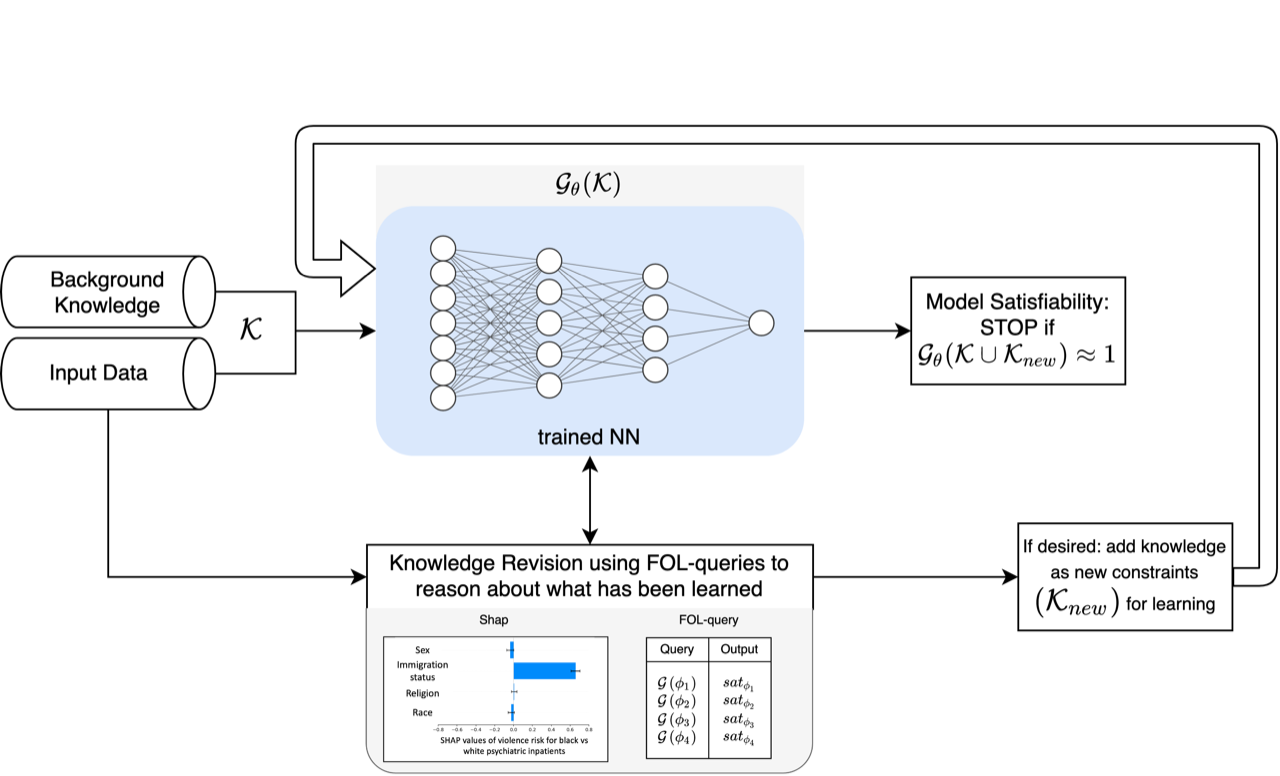}
\caption{%SHAP values for each feature. 
Illustration of the LTN pipeline for continual interactive learning: revision is carried out by querying a deep network interactively and learning continually, thus applying the neural-symbolic cycle multiple times. Explanations extracted from the network using, e.g., SHAP can highlight bias or undesired properties in feature importance. Equally, querying the network in LTN-style shows the satisfiability of specific model properties, such as fairness constraints, which can subsequently be added to the knowledge base $\mathcal{K}$ for further training. Doing this, we can answer questions such as: \textit{How does the model behave for a specific group of individuals compared to others?}, by translating into FOL queries %:  $\forall x \in \mathcal{R}_{P1}, y \in \mathcal{R}_{U1}: D(x) \leftrightarrow D(y)$ 
and checking their degree of satisfiability (\textit{sat()}) (c.f. Section \ref{section:nesycasestudy}). %\nat{notation is not introduced yet and thus these formulas still say little to me. What is D, P1 and U1? are subindexes necessary? what is R?}
Subsequently, such desired queries can be added to the optimisation function. This process concludes once it has been shown to reduce bias at a subsequent SHAP explanation.} %useful post, and another of Lundberg (there is another on demographi parity): https://towardsdatascience.com/be-careful-when-interpreting-predictive-models-in-search-of-causal-insights-e68626e664b6
%\nat{it would be great if you can put an example of query in nat language and in LTN syntax here and also in the figure, as it is hard to imagine how a FOL query or satisfiability degree of a concept is converted into a constraint for revised knowledge. Also indicate 'where Sat() stands for satisfiability in FOL (can you give more details of the particular logic? } 
\label{fig:nesy}%\nat{mention in caption what the shap model is applied on: recicivism prediction? etc}
\end{figure}

\subsection{Framework for interactive explainability}

In a more tightly integrated Neural-Symbolic system, XAI occurs in line with the neural-symbolic cycle. In this framework, we can query and revise the information and consolidate existing background knowledge. The system can utilize background knowledge to provide meaningful semantics for the explanations, facilitating human-machine interaction and achieving desired properties. 

By applying the neural-symbolic cycle multiple times, partial symbolic descriptions of the knowledge encoded in the deep network can be checked and, through a human-in-the-loop approach \cite{Holzinger:2016:iML}, incorporated into the cycle as a constraint on the learning process. This enables an interactive integration of a desired behaviour, notably fairness constraints, by verifying and incorporating knowledge at each cycle, instead of (global or local) XAI serving only to produce a one-off description of a static system.

The neural-symbolic cycle can be seen as a common ground for communication and system interaction, allowing for a human-in-the-loop continual learning approach. This enables an interactive integration of a desired behaviour, notably fairness constraints, by verifying and incorporating knowledge at each cycle, instead of (global or local) XAI serving only to produce a one-off description of a static system. %\nat{use consistently neural-symbolic thorough the article} todo cite lesortLomonacoDiaz
Symbolic knowledge representation extracted from the learning system at an adequate abstract level for communication with users should allow knowledge consolidation and targeted revision. 

The following example will demonstrate how to use the Logic Tensor Networks (LTN) framework for explainable classification tasks and subsequently address some undesired model properties according to the pipeline in Figure \ref{fig:nesy}. %\nat{doesnt seem like the pipeline is described inmediately? check or refer to pipeline in fig 9} 
In it, we use the Shapley method, but any other XAI method could have been chosen, and an integrated logical Neural Network querying mechanism, to gain insights into the model during our knowledge revision process.

\subsection{Logic Tensor Networks (LTN) for explainable model revision}%\nat{to generate NLE? be more explicit on the purpose wrt XAI of each method presented in each subsection}}
The framework used in this approach and accompanying notebook\textsuperscript{\ref{foot:nesylink}} %\nat{can we say notebook? will we have one?} 
is LTN \cite{Serafini2016,serafini2021logic,Badreddine2020}. However, instead of treating the learning of the parameters from data and knowledge as a single process, we emphasise the dynamic and flexible nature of the process of training from data, querying the trained model for knowledge, and adding knowledge in the form of constraints for further training, as part of a cycle whose stopping criteria are governed by a fairness metric. Furthermore, we focus on the core of the LTN approach: con\-straint-based learning from data and first order logic knowledge (FOL). We make the explanation approach iterative by saving the learned parametrisation at each cycle in our implementation, while changing the original LTN implementation from Neural Tensor Networks %\nat{new term not introduced before? not sure I get this long sentence, maybe 'changing the original LTN implementation using NTN?'}
to a feed-forward Neural Network to demonstrate the agnostic nature of the approach.  %\nat{last sentence is unclear, using NTN to a FFNN? what does it mean?}

Whereas many inherently neural-symbolic methods come with stringent architectural constraints on the model itself, this LTN adaptation is model-agnostic since LTN as a framework solely requires the ability to query any deep network (or any ML model) for its behaviour, that is, observing the value of an output given a predefined input. The predictive model itself can be chosen independently, with the LTN acting as an interface to provide the explanation of the model to the user in the form of targeted FOL queries. 

Logic Tensor Networks \cite{Serafini2016,Badreddine2020} implement a many-valued FOL language $\mathcal{L}$, which consists of a set of constants $\mathcal{C}$, variables $\mathcal{X}$, function symbols $\mathcal{F}$ and predicate symbols $\mathcal{P}$. 
Logical formulas in $\mathcal{L}$ allow to specify background knowledge related to the task at hand. 
The syntax in LTN is that of FOL, with formulas consisting of predicate symbols and connectives for negations ($\neg$), conjunction, disjunction and implication ($\land,\lor,\rightarrow$) and quantifiers ($\forall,\exists$).

\textbf{Learning in the LTN framework for explanation}: LTN functions and predicates are learnable. Thus, the grounding of symbols depends on a set of parameters $\theta$. With a choice of a multilayer perceptron as model, each logical predicate is represented by a feed-forward mapping, where $\sigma$ denotes the sigmoid activation function which ensures that predicate $P$ is mapped from $\mathbb{R}^{mxn}$ to a truth-value in $[0,1]$.\\
Since the grounding of a formula $\mathcal{G}_\theta(\phi)$ denotes the degree of truth of $\phi$, one direct training signal is the degree of truth of the aggregate of the formulas %\nat{what are aggregate formulas? maybe specify rapidly or in footnote?}
in the knowledge base $\mathcal{K}$. The aggregate of the entries in $\mathcal{K}$ may be achieved by averaging all terms using the mean but alternative approaches exist \cite{Badreddine2020}. 
% The general mean is used because it gives flexibility to the user to tweak %\nat{how does the user determine this?does he input any parameter in this formula?  what is the strictness of the aggregation?} 
% the method of the aggregation (by specifying \cnat{minimum probability of satisfaction? name it} $p$), meaning the relative importance of smaller and larger values\footnote{\label{foot:pmean} $\textit{p-mean}(x_1,\dots,x_n) = \biggl( \frac{1}{n} \sum\limits_{i=1}^n x_i^p \biggr)^{\frac{1}{p}}$.}. 
The objective function $\mathrm{Sat}_{A}(\mathcal{G}_\theta(\mathcal{K}))$ %\nat{noted how? with which letter? is a simple loss function of a DNN?}
is therefore the satisfiability of all formulas in $\mathcal{K}$ which are maximized by tweaking the model parameters. 
%$\theta$:
%$$
% \theta^\ast = \argmax_{\theta\in\Theta} \mathrm{Sat}_{A}(\mathcal{G}_\theta(\mathcal{K}))
%$$
%which are subject to an aggregation $A$ of all formulas, e.g. the generalised mean (p-mean). % \nat{of all axioms' satisfiability}. %\textsuperscript{\ref{foot:pmean}}. 
In its simplest form of binary classification without constraints, $\mathcal{K}$ will consist of one term for positive examples in the dataset and one for negatives.  
Notice that the approach described in Figure \ref{fig:nesy} is model-agnostic. The core extension of regular neural network optimisation enabled by LTN is that of querying with many-valued first order logic and learning with knowledge base constraints. %\nat{keep only one formula per p-mean, only first time its used, not 4 for same one}

\textbf{Continuous Querying for model understanding}: LTN inference using first order logic clauses is not only a post-hoc explanation in the traditional sense. It allows that inference forms an integral part of an iterative process allowing for incremental explanation through the distillation of knowledge guided by data.
We achieve this by computing the value of a grounding $\mathcal{G}_\theta(\phi_q)$, given a trained network (set of parameters $\theta$), for a user-defined query $\phi_q$.

Specifically, we save and reinstate the learned parameters stored in the LTN implementation. This is done by storing the parameters $\theta$ resulting from %$\theta^\ast =  \argmax_{\theta\in\Theta}\mathrm{Sat}_{A}\mathcal{G}_\theta(\mathcal{
%K})$. This also means that changes to the knowledge base will instead start from saved state $\theta^\ast$. Such mechanism allows us to continually query and guide the learning process according to the added knowledge $\mathcal{K}_{new}$, an approach akin to that of continual learning.\\% (Algorithm \hyperlink{fig:ltnalgo}{2}). \\
% \nat{make sure the algo LTN-active learning cycle goes here? I dont see a way to reference it/dont find it in the .tex}
% \nat{add proper ref to the algorithm with alg:X and number each algo line to facilitate feedback. It also says Algorithm 1 when it should be algo 2 (see gradcam algo)}
A query is any logical formula expressed in first order logic. Queries are evaluated by calculating the grounding $\mathcal{G}$ of any formula whose predicates are already grounded in the multilayer perceptron or even by defining a predicate in terms of existing predicates. For example, the logical formula $\forall x: (A(x) \rightarrow B(x))$ can be evaluated by applying the values of $x$, obtained from the dataset, to the trained Neural Network, obtaining the values of output neurons $A$ and $B$ in [0,1] (corresponding to the truth-values of predicates $A$ and $B$, respectively), and calculating the implication with the use of the Reichenbach-norm and aggregating for all $x$ using the p-mean. %\textsuperscript{\ref{foot:pmean}}. 

A query can explain AI systems by connecting different model outputs, aggregating inputs for summarising the behaviour of a system in specific domains or relating specific inputs with specific features against each other and the output. In \cite{wagner2022neuralsymbolic}, the authors demonstrate that this framework can be extended to include intermediate representations, thus providing an understanding of how concepts are logically associated with particular output classes.

Logical formulas used for such explanations follow semantics for logical connectives that are defined according to fuzzy logic semantics: conjunctions are approximated by t-norms (e.g., $min(a,b)$), disjunctions by t-conorms (e.g. $max(a,b)$), negation by fuzzy negation (e.g. $1-a$) and implication by fuzzy implications (e.g. $max(1-a,b)$). 
Semantics of quantifiers are defined by aggregation functions. 
In the following example, we approximate the binary connectives using the product t-norm and the corresponding t-conorm which define our fuzzy logic. The universal quantifier is defined as the generalised mean, also referred to as p-mean. %\textsuperscript{\ref{foot:pmean}}.%\nat{p mean is defined here but used much earlier, twice and with no clear defineiton, move definition to first time it is mentioned in page 13}

% \nat{please add correct hyperlink to algo 2 and full source code here, to modify it if needer homogenize to dataset, not data-set} BW: should be fixed now - would need to rewrite the entire algorithm for the package used in the previous section (as we wanted comments in line). Can do so if harmonisation is desired here
\begin{figure}[h!]
     \centering
     \includegraphics[width=0.45\textwidth]{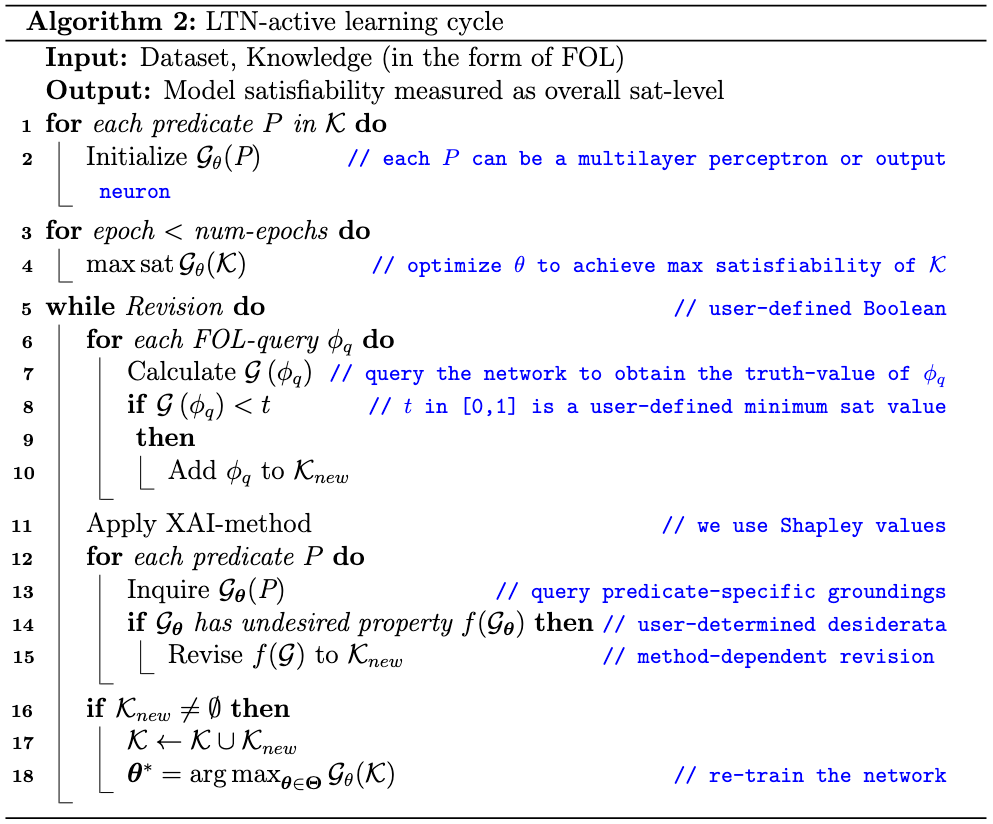}
     \hypertarget{fig:ltnalgo}{}
     % \label{fig:ltnalgo}
 \end{figure}
 Algorithm \hyperlink{fig:ltnalgo}{2} illustrates the steps we take to continuously refine $\mathcal{K}_{new}$ with a human-in-the-loop. 
The queries derive from questions a user might have about the model's response: \textit{How does the model behave for a specific group? How does the model behave for particular edge cases?} These questions can be translated into FOL-queries. Simultaneously, an XAI method further informs the user about possible undesired model behaviour which may not be as apparent as the above common questions
In Figure \ref{fig:nesy}, XAI method SHAP reports a disparity in how the variable \emph{immigrant status} is used by the ML system for black % and women 
and white inpatients when predicting %setting credit risk \cite{Wagner2021}
the risk of violence in a psychiatric hospitalization setting \cite{sikstrom2022conceptualising}. 

%IVan new case study is recidivism of violence for inpatients in psychiatric settings,  that is, whether mad people tend to reiterate violence during an hospitalization.   you can remove recidivism of violence and substitute with readmission.  The idea is: you need xai models for improving the fairness.  You can't predict that black inpatients people will be more likely to violence recedivism (This is a bias due to some feats). - This is the problem they want to solve and they did this with a pipeline SHAP + LTN (well, they did it for another domain, I just adapted it.)
The XAI technique SHAP is used together with LTN queries to highlight such findings and subsequently address them by adding knowledge to $\mathcal{K}_{new}$ and retraining, as will be illustrated in the next section and the accompanying notebook.%\footnote{\ref{foot:nesylink}}.  
%\cnat{@Adrien fix footnote to include COMPAS (Correctional Offender Management Profiling for Alternative Sanctions), dataset used to assess a convicted criminal’s likelihood of reoffending, available in \cite{larson2016data} \url{https://github.com/propublica/compas-analysis}}  
This cycle can be repeated until the revision process delivers satisfactory results to the user, with respect to model performance and behaviour.

\subsection{Case Study on % recidivism prediction 
violence reiteration prediction in a psychiatric hospitalization setting with a Deep %\nat{Deep? or a MLP? clarify }
Neural Network within the LTN framework for explainable revision} \label{section:nesycasestudy} %\nat{on appplication X with a X model (classif, prediction, regression}, lessons learned and next steps \nat{include applying X on model Y through XAI method Z}}

We demonstrate the method mentioned above using a well-known fairness-related \emph{COMPAS} dataset, % obtained from the UCI machine learning repository 
from ProPublica\footnote{\label{foot:nesylink}The demonstration of the method, the XAI + LTN pipeline (on which XAI techniques such as SHAP can be embedded for interactive explanations) and the data are accessible at \url{https://colab.research.google.com/drive/1Ip9Yb9gVRSRqaBKY9gOpiWn9pq3LovWG?usp=sharing}. The original LTN repository adapted for this method is: \url{https://github.com/logictensornetworks/logictensornetworks}}. For additional examples using alternative datasets, as well as comparisons with alternative methods, we refer the reader to \cite{Wagner2021}. % ToDo new adaptation: 
We use a case study of violence risk prediction of inpatients in a psychiatric hospitalization, i.e., to predict whether inpatients tend to reiterate violent behaviour or not during their stay % hospitalization 
\cite{sikstrom2022conceptualising}\footnote{\url{https://informatics.bmj.com/content/bmjhci/29/1/e100459/DC1/embed/inline-supplementary-material-1.pdf?download=true}}. %  (The new case study is recidivism of violence for inpatients in psychiatric settings. We can remove recidivism of violence and substitute with readmission, and reinsertion in the hospital or readmission instead of recidivism.
%new use case:
Typical features of this case study encode sociodemographic information such as sex, race, immigration status, etc. ML algorithms tend to generate false positives in presence of a protected group versus an unprotected group (black vs white, immigrants vs non immigrants). In this case, a bias on such features may compromise fairness. The presented framework (based on XAI (SHAP)+LTN) shows how such \textit{MLOps} pipeline can detect it and tackle it.
%In this case, the fairness is compromised by features such as the race (the protected variable) or immigrant status.

%We train a neural network with two hidden layers of 100 and 50 neurons, respectively. We use the Adam optimiser with a learning rate of 0.001 trained for a maximum of 5000 epochs. Then $\forall x \in \mathcal{T}_R: D(x)$ and 
%$\forall x \in \mathcal{T}_N: \neg D(x)$, where $\mathcal{T}_R$ is a set of individuals who re-offend; 
%and $\mathcal{T}_N$, a set of individuals who do not. Hence, we consider recidivism prediction as a traditional supervised learning setting.

%We subsequently query and revise the model on race-related disparities in the COMPAS dataset. 

%The
A trained network is queried to return the truth value associated to the predicate used for the classification task $\mathcal{G}(D(\mathcal{T}))$ for the entire training set $\mathcal{T}$. 
This will allow us to answer how the model treats similar individuals across protected and unprotected groups.
% \nat{next sentence is too long and not understandable, try simplify/split: (use maybe differ instead of different? avoid long sentence}
Using quantile-based discretisation, we obtain answers to the question: \textit{How prediction for equally sized groups for each protected and unprotected variables differ across different risk categories for %reoffending
violence?} 
%$\forall x \in \mathcal{R}_{P1}, y \in \mathcal{R}_{U1}: D(x) \leftrightarrow D(y)$, where $\mathcal{R}_{P1}$ corresponds to the different individuals of risk categories $i$ %\nat{you mean feature values?} 
%of a protected group and $\mathcal{R}_{U1}$ of an unprotected group. 
% We find out whether the model achieves material equivalence %\nat{what is aggregated material equivalence?} 
% between black and white prisoners of the same risk category on aggregation.  %\nat{can you rephrase what this if and only if means in this context? teh amount of male and female recidivism is the same? if and only if? make the expresssion read in natural nlangauge and what R and D mean here. It is not obvious what this is trying to say} 
% Querying such axioms reveals a low satisfiability level ($sat_{\phi_{i}}\approx0.5$), %\nat{of concept phi n? why not concept phi j or phi i, clarify how phi n relates to P1 and U1 } ,
% showing that the model is picking up on significant disparities in groups of medium risk of recidivism. This means that when the model is uncertain, it will predict re-offending for the protected group more often than for the unprotected group. % \nat{is this a bad /unfair thing? it is not clear from this example, please clarify}  
We determine whether the model achieves parity between black and white prisoners in the same risk category on aggregate.  
Querying such axioms reveals a low level of satisfiability ($sat_{\phi_{i}}\approx0.5$), suggesting that the model is learning undesirable disparities in groups with medium risk of %recidivism
violence. Thus, the model predicts bad behavior more often for the protected group than for the unprotected group of the same risk category. 
In groups where the risk of %recidivism
violence is very low or very high, there are no significant differences in violence prediction of the model %\nat{what does it mean treatment in recidivism prediction application? please briefly contextualize. Could you please mention the fairness framework to use to fix that undesired disparity that you use? there is plenty in ML blooming.} 
for the protected and unprotected group, which is in line with the desired notion of \textit{group fairness} \cite{Dwork2012}.

We confirm the disparity between groups by calculating their Shapley values. Since the SHAP method uses the same units as the original model output, we can decompose the model output using SHAP and compute each feature's parity difference among protected and unprotected groups using their respective Shapley value.

We can subsequently revise $\mathcal{K}$ using the queries $\phi_{i}$ %\nat{what are truth queries? queries over the ground truth dataset you mean?}
of the different groups as soft constraints and are able to revise the network to decrease the undesired disparities while retaining high accuracy as measured in \cite{Wagner2021} and the notebook\textsuperscript{\ref{foot:nesylink}}. 
%\nat{can you have a notebook further exemplifying this to show how much accuracy drops and how much fairer is the decision (ie. through shap analysis? after iterative explanation querying and model fixing? } 
In this demonstration, only model outputs combined with protected attributes are used to inform the queries, as the focus is primarily querying the output concerning unfair treatment. A further query could answer how predictions differ across groups of a specific age in combination with protected attributes. 

Any combination of features or even intermediate representations, such as feature activations in a CNN, as well as a combination of models are available and can be queried using the LTN framework through fuzzy logic. Furthermore, the latest iteration of the LTN framework allows for dynamic masking, which means that the explanation iterations and revision could be further automated within the LTN framework using custom masks. Such custom masks, for example, remove the necessity of manual discretisation into parity groups by performing automatic grouping based on dynamically changing output logits. 

In our example, however, the user can vary the number of user-defined queries and discretisations groups into different granularities. It is worth emphasizing the flexibility of such approach w.r.t. further queries and its potential use with alternative fairness constraint constructions. With the increasing complexity of models as well as fairness definitions, rich languages such as fuzzy FOL can be beneficial to adapt to regulatory and societal changes to notions of fairness. One example would be a simple adaptation of the value $p$ in the aggregation using the p-mean. Using larger values for $p$, the fairness notion converges from \textit{group fairness} towards \textit{individual fairness}, as the generalised mean, converging from a simple average towards the $min$ value, gradually (with higher relative importance for lower values).

Integrating XAI methods with neural-symbolic approaches allows us to learn about the undesired behaviour of a model and intervene to address discrepancies, which is ultimately an important goal of explainability. We have demonstrated an interactive model-agnostic method and an algorithm for fairness in healthcare and have shown how one can remove demographic disparities from trained neural networks by using a continual learning LTN-based framework.

%%%%%%%%%%%%%%%%%%%%%%%%%%%%%%%%%%%%%%%%%%%%%%%%%%%%%%%%%%%%%%%%%%%%%%%%%%%%%%%%%%%%%%%%%%%%%%%%%
\section{Rendering XAI explanations through a template system for natural language explanations (TS4NLE)} \label{NLG}
%AI model explanations through natural language
%Generating AI model explanations through natural language
%Graph-based explanations: Tailored natural language explanations through templates

All methods discussed above provide outputs in a structured format that can be represented in a graph-like way.
Such a format enables the design of different strategies for transforming the provided outputs into a representation that can be easily understand and consumed by the target user.

Explanations generated starting from structured formats such as the one mentioned above help users in better understanding the output of an AI system.
A better understanding of this output allows users to increase the overall acceptability in the system.
An explanation should not only be correct (i.e. mirroring the conceptual meaning of the output to explain), but also useful.
An explanation is useful or actionable if and only if it is meaningful for the users targeted by the explanation and provides the rationale behind the output of the AI system~\cite{WhatDoesExplainableAImean}. %\cite{arrieta2020explainable}
%Explanations are meaningful if they are easily understandable by the targeted audience depending on the context in which the explanations are received.\nat{sounds a bit redundant, may be cut if this is moved to intro and/or discussion}
For example, if an explanation has to be provided on a specific device, such a device represents a constraint to be taken into account for deciding which is the most effective way for generating the explanation. Such explanation can be in natural language/vocal messages, visual diagrams or even haptic feedback.

In this Section, we focus on the generation of Natural Language Explanations (NLE). Producing these carries a challenge, given the requirement of adopting proper language with respect to the targeted audience \cite{arrieta2020explainable} and their context.
Briefly, let us consider a sample scenario occurring within the healthcare domain where patients suffering from diabetes are monitored by a virtual coaching system in charge of providing recommendations about healthy behaviors (i.e. diet and physical activities) based on what patients ate and which activities they did.
The virtual coaching system interacts with both clinicians and patients, and when an undesired behavior is detected, it has to generate two different explanations: one for the clinician containing medical information linked with the detected undesired behavior --including also possible severe adverse consequences; and one for the patient omitting some medical details and, possibly, including persuasive text inviting to correct the patient's behavior in the future.
% as universal medium for communication.
%\subsection{Tailored natural language explanations through templates}
The end-to-end explanation generation process, from model output to an object usable by the target users, requires a building block in the middle supporting the rendering activity.
Such rendering requires explanations having a formal representation with a logical language equipped with predicates for entities and relations. This formal representation can be directly represented as an \emph{explanation graph} with entities/nodes and relations/arcs. It and allows: i) its own enhancement with other concepts from domain ontologies or Semantic Web resources; and, ii) an easy rendering in many human-comprehensible formats. 
Such an explanation graph can be easily obtained from the XAI techniques explained above. 
For example, the explanatory features and the output class provided by SHAP can be regarded as the nodes of the explanation graph, whereas arcs are computed on the basis of the SHAP features values.
SHAP's output is one of the possible inputs that the TS4NLE strategy can process.
Indeed, TS4NLE is agnostic with respect to the type of model adopted by the ML system, since it can work with any approach providing an output that can be represented with a graph-like format.
The \emph{explanation graph} can also work as bridge for accessing different types of knowledge usable, for example, to enrich the content of natural language explanations by respecting privacy and ethical aspects connected with the knowledge to use.

\begin{figure*}[h]
\centering
\includegraphics[scale=0.34]{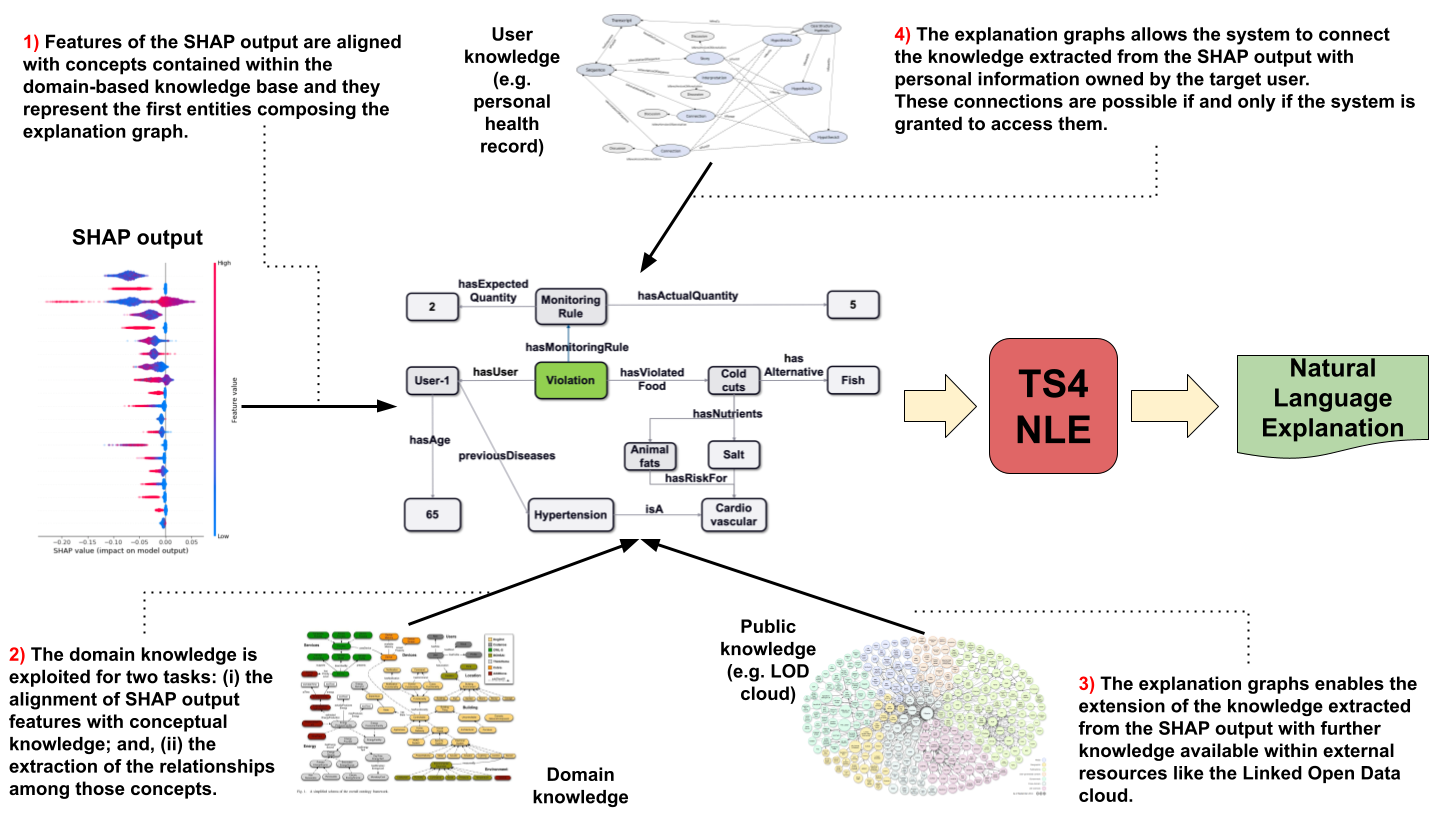}%\nat{it would be great to show the names of the shap features analized in shap plot on the left if space can be ade so that it is readable, also name the domain from where each domain knowledge graph (LOD, KG, User knowledge graph are about/named)}
\caption{The images show the process of transforming a SHAP output into an explanation graph that is then transformed into its equivalent natural language explanation. Features contained within the SHAP output are transformed into concepts linked with a knowledge base related to the problem's domain. Such a knowledge base is exploited also for extracting relationships between the detected concepts. This preliminary explanation graph can be enriched with further knowledge extracted from publicly available resources (e.g. the Linked Open Data cloud) as well as with private data (e.g. personal health records). Finally, the explanation graph, through the NLE rendering component is transformed into a natural language explanation.}%\nat{cite paper where this is published if not novel, in the caption.}
\label{fig:graph_expla_overall} 
\end{figure*}

Explanations require a starting formal (graph-like) representation to be easily rendered and personalized through natural language text~\cite{Donadello19}. 
%Such an explanation graph can be obtained by the standard XAI techniques explained above. 
The generation of such natural language explanations can rely on pipelines which takes the structured explanation content as input and, through several steps, performs its linguistic realization%~\cite{ReiterD97}.
The work in~\cite{Donadello19} injects in such a pipeline a template system that implements the text structuring phase of the pipeline. 
Figure~\ref{fig:graph_expla_overall} shows the explanation generation process starting from a SHAP analysis of a model output.

As mentioned above, generating natural language explanations starts from the creation of the explanation graph, since it provides a complete structured representation of the knowledge that has to be transferred to the target user.
As first step, the features of the SHAP output are transformed into concepts of the explanation graph and they are, possibly, aligned with entities contained within the knowledge base related to the problem's domain.
Such entities represent the first elements composing the explanation graph that can be used as collector for further knowledge exploited for creating the complete message.
Beside the alignment of SHAP output features with the domain knowledge, such a knowledge base is exploited for extracting the relationships among the identified concepts.
The extraction of such relationships is fundamental for completing the explanation graph as well as for supporting its transformation into its equivalent natural language representation.
Once the alignment between the SHAP output and the domain knowledge has been completed, the preliminary explanation graph can be extended in two ways.
First, public available knowledge can be linked to the preliminary explanation graph for completing the domain knowledge. Let us consider as example the explanation graph shown in Figure~\ref{fig:graph_expla}. Some medical information associated with the identified food category may not be contained in the domain knowledge integrated into the local system. Hence, by starting from the concept representing the food category, we may access, through the Linked Open Data cloud, the UMLS\footnote{Unified Medical Language System (UMLS) \url{https://www.nlm.nih.gov/research/umls/index.html}} knowledge base for extracting information about the nutritional disease risks connected with such a food category.
Beside public knowledge, the explanation graph can be enriched with user information provided if and only if they are compliant with respect to possible privacy constraints.
User information can be provided by knowledge bases as well as probabilistic models. Also in this case TS4NLE is agnostic with respect to the external source to exploit.
In the use case we present below, TS4NLE relies on an external user-oriented knowledge base containing facts that TS4NLE can reason on for deciding which kind of linguistic strategy to adopt.
Let us consider the healthcare domain use case.
Here, information contained in the users' personal health record can be used for enriching the explanation graph with concepts by linking, for example, the negative effects of the over-consumption of a specific food category %with possible user's nutritional diseases.
by users with potential nutritional diseases.
% Finally, the explanation graph is given as input to the TS4NLE module that is in charge of performing all reasoning and rendering operations with the aim of generating the proper natural language explanation.

%\nat{typo in upper text: The explanation graphs allows- allow. What is the matching algo used? name the center graph, OWL ontology, KG, etc, for each graph shown. Put also in parenthesis what the expalnation graph is an ont or a KG? it is not obvious what is a KG or an explanation graph and what is the difference.   a box representing this process is missing? A LOD cloud can be an explanation? make more clear what are examples of explanation graph and what is each. Uppermost graph is not readable, Could it be possible to reduce arrows space or reorganize text around it to make it bigger?}
%\nat{surround this fig by text-so it doesnt waste a full page}
%\nat{fig 5: where is the shap features and graph entities matching happening? seems like an entity resolution process module is missing in this pipeline to match shap features with graph entities/nodes (a explanation graph cannot do it alone Am I right?}

Finally, the created explanation graph can be rendered in a natural language form through a template system for natural language explanations (TS4NLE)~\cite{Donadello19} that leverages a Natural Language Generation (NLG) pipeline. 
Templates are formal grammars whose terminal symbols are a mixture of terms/data taken from the nodes/arcs of the explanation graph and from a domain knowledge base. 
Terms in the explanation graph encode the rationale behind the AI system decision, whereas the domain knowledge base encodes further terms that help the user's comprehension by:
i) enhancing the final rendered explanation with further information about the output; and, ii) using terms or arguments that are tailored to that particular user and increase the comprehension of the explanation. 
Generally, this user model is previously given, in form of an ontology or knowledge graph.

TS4NLE is structured as a decision tree where the first level contains high-level and generic templates that are progressively specialized and enriched according to the user's feature specified in the user model. 
Once templates are filled with non-terminal terms, the lexicalization\footnote{\textit{Lexicalization} is the process of choosing the right words (nouns, verbs, adjectives and adverbs) that are required to express the information in the generated text, it is extremely important in NLG systems that produce texts in multiple languages. Thus, the template system chooses the right words for an explanation, making it tailored.}
and linguistic realization of the pipeline are performed with standard natural language processing engines such as RosaeNLG\footnote{\url{https://rosaenlg.org/rosaenlg/3.0.0/index.html}}.

\subsection{%Natural language explanations use case: Persuasive messages for healthy lifestyle adherence}
TS4NLE use case: persuasive message generation for healthy lifestyle adherence}

In this subsection, we provide the description of a complete use case related to the generation of persuasive natural language explanation within the healthcare domain.

Given as input a user lifestyle (obtained with a diet diary or a physical activity tracker), AI systems are able to classify the user behavior in classes ranging from \textit{very good} to \textit{very bad}. The explanation graph contains the reason for such a prediction and suggestions for reinforcing or changing the particular lifestyle. According to the user model (e.g., whether the user has to be encouraged or not, the users' barriers or capacities), the template system is explored in order to reach a leaf containing the right terms to fill the initial non-terminal symbols of the template. 
A user study regarding the Mediterranean diet states that such tailored explanations are more effective at changing users' lifestyle with respect to a standard notification of a bad lifestyle. %for long version \cite{Donadello19}
A further guide of this use case is available online\footnote{\url{https://horus-ai.fbk.eu/semex4all/}}.

\begin{figure}[htbp!]
\centering
\includegraphics[scale=0.28]{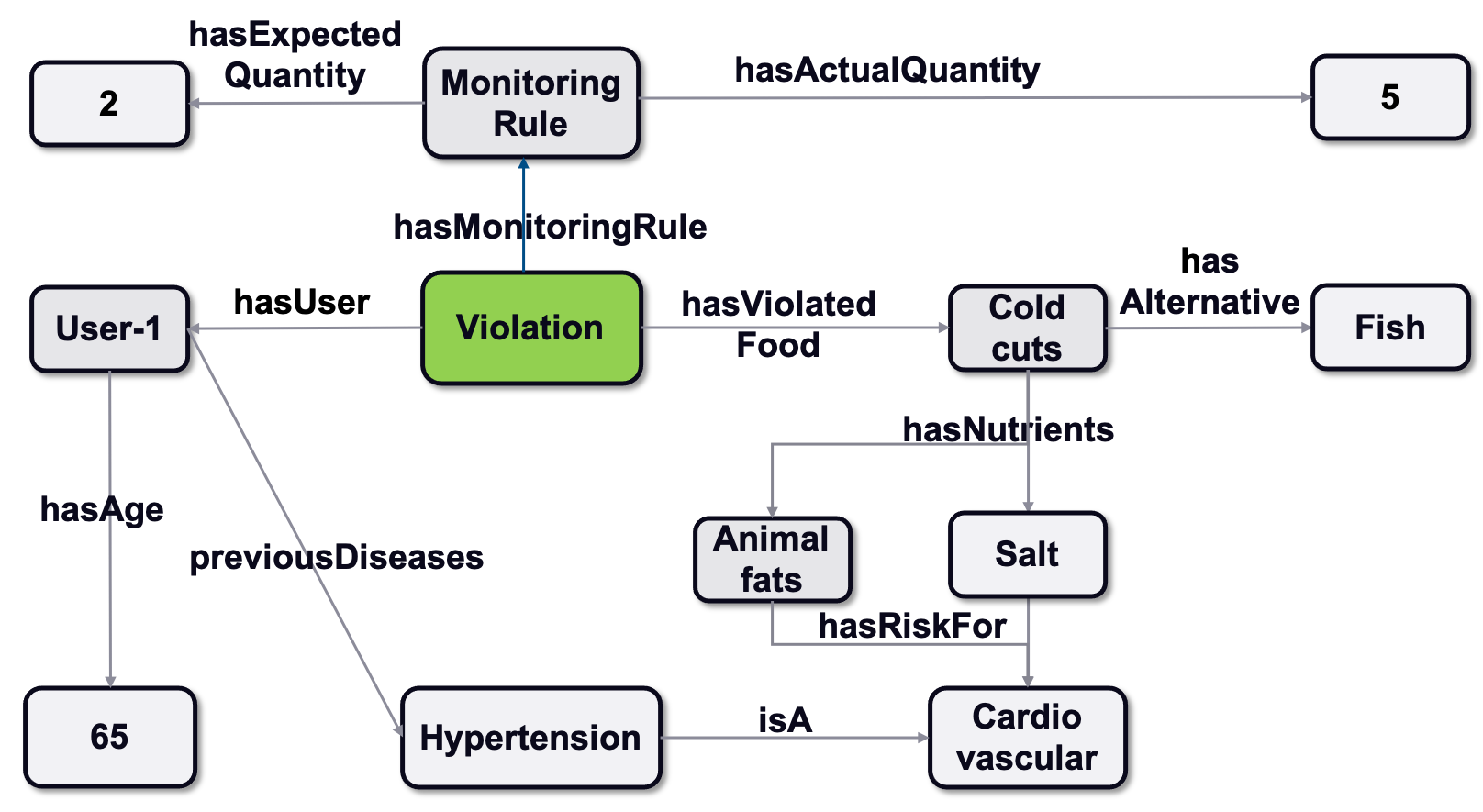}
\caption{Explanation graph for users exceeding in cold cuts consumption in the diet \& healthy lifestyle adherence application.}
\label{fig:graph_expla}
\end{figure}

%\subsubsection{TS4NLE visualization}
The explanation graph contains entities connected by relations encoding the rationale of the AI system decision. Fig. \ref{fig:graph_expla} contains the explanation graph for a 65 years old user that consumes too much cold cuts.
Such a graph is rendered with TS4NLE as: \textit{``This week you consumed too much (5 portions of a maximum 2) cold cuts. Cold cuts contain animal fats and salt that can cause cardiovascular diseases. People over 60 years old are particularly at risk. Next time try with some fresh fish''}.

%\subsection{TS4NLE running example}
The generation of the natural language explanation shown above is performed by TS4NLE by following the steps below.
After the generation of the explanation graph, the \textit{message composition} component of TS4NLE starts the generation of three textual messages for the feedback, the argument and the suggestion, respectively.
This is inspired by the work in~\cite{denAkker2015} and expanded taking into consideration additional strategies presented in~\cite{guerini:et:al:AAI-07}. %\nat{missing ref}
These consist of several persuasion strategies that can be combined together to form a complex message. Each strategy is rendered through natural language text with a template. A template is formalized as a grammar whose terminal symbols are filled according to the data in the violation package and new information queried in the reference ontology. Once templates are filled, a sentence realizer (i.e. a producer of sentences from syntax or logical forms)%~\cite{GattR09}
generates natural language sentences that respect the grammatical rules of a desired language

%\footnote{Current version of TS4NLE supports the generation of messages in English and Italian. In particular, Italian language requires a morphological engine (based on the open-source tool called morph-it\footnote{\url{https://docs.sslmit.unibo.it/doku.php?id=resources:morph-it}}) 
%to generate well-formed sentences starting from the constraints written in the template (e.g., tenses and subject consistency for verbs)}. 

Below we describe the implemented strategies to automate the message generation, focusing also on linguistic choices.
%The template model together with an example for instantiating it, is represented in Figure~\ref{arg_message}.

\par \textbf{Explanation Feedback}:
%\footnote{The feedback concept in the message generation model of~\cite{denAkker2015} must not be confused with the behavior change strategy element \textit{feedback} in the BIT model.}:
is the part of the message that informs the user about the not compliant behaviour, hereafter called ``violation'', with the goal that has been set up.
Feedback is generated considering data included in the explanation graph starting from the violation object: the food entity of the violation will represent the object of the feedback, whereas the level of violation (e.g., deviation between food quantity expected and that actually taken by the user) is used to represent the severity of the incorrect behavior. 
The intention of the violation represents the fact that the user has consumed too much or not enough amount of a food entity. Feedback contains also information about the kind of meal (breakfast, lunch, dinner or snack) to inform the user about the time span in which the violation was committed. 
%commented out due to the figs limitation
%The template aligned with the terminal symbols of the violation in our running example is in Figure \ref{feed_message}.
%\begin{figure}[!htpb]
%\centering
%\includegraphics[width=0.5\textwidth]{figures/feed_message}
%\caption{TS4NLE model (template and example of violation) for generating the text of the feedback. Choices on template and message chunks depend on the violation package. This holds also for both the argument and suggestion. Dashed lines represent a dependency relation. A template of type ``informative'' has been used in the example.}
%\label{feed_message}%\nat{clarify in figs 7 and 8 the type of template used in the caption}
%\end{figure}
%ToDo can be prescinded if space needed
From a linguistic point of view, choices in the feedback type are related to the verb and its tense: e.g., beverages imply use of the verb \textit{to drink} while for solid food we use \textit{to eat}. To increase the variety of the message, verbs \textit{to consume} and \textit{to intake} are also used. Past simple tense is used when violation is related to a specific moment (e.g. \textit{You drank a lot of fruit juice for lunch}), while present continuous is used when the violation is related to a period of time of more days and the period is not yet ended (e.g., \textit{You are drinking a lot of fruit juice this week}). 

\par \textbf{%\nat{explanation argument? } 
Explanation Argument}: it is the part of the message informing users about possible consequences of a behavior.
For example, in the case of diet recommendations, the \textit{Argument} consists of two parts: (i) information about nutrients contained in the food intake that caused the violation and (ii) information about consequences that nutrients have on human body and health. Consequences imply the positive or negative aspects of nutrients. %The template for this BCS is shown in Figure~\ref{arg_message}.
%\begin{figure}[htbp!]
%\centering
%\includegraphics[width=0.5\textwidth]{figures/arg_message}
%\includegraphics[scale=0.40]{figures/HORUS-template-full.png}
%\caption{
%TS4NLE model (template and example of violation) for generating the text of the argument given as part of the explanation when violating diet restrictions.
%TS4NLE model (template and example of violation) for generating the text of an explanation. The top part represents the generation of the feedback. Choices on template and message chunks depend on the violation package. This holds also for both the argument and suggestion. Dashed lines represent a dependency relation. A template of type ``informative'' is used in the example. The middle part represents the generation of the argument, which is given as part of the explanation when violating diet restrictions. Finally, the bottom part represents the generation of the suggestion.
%}
%\label{arg_message}
%\end{figure}
In this case, TS4NLE uses the intention element contained in the selected violation package to identify the type of argument to generate. Let us consider the violation of our running example where the monitoring rule limits the daily fruit juice drinking to less than 200 ml (a water glass) since it contains too much sugar. In the presence of an excess in juice consumption (translating to a discouraging intention) the argument is constituted by a statement with the negative consequences of this behavior on user health. On the contrary, the violation of a rule requiring the consumption of at least 200 gr of vegetables per day brings the system to generate an argument explaining the many advantages of getting nutrients contained in that food (an encouraging intention). In both cases, this information is stored within the explanation graph. 
Moreover, TS4NLE analyzes the message history to decide which property of the explanation graph to use in the \textit{Argument}, to generate a message content that depends on e.g., content sent in the past few days, ensuring a certain degree of variability. With respect to linguistic choices, the type of nutrients and their consequences influence the verb usage in the text. Finally, to emphasize different aspects of the detected violation, templates encode the use of appropriate parts of speech. For example, for stressing the negative aspects of the violated food constraint, the verb \textit{contain} (nutrients) and \textit{can cause} (for consequences) were used. On the other hand, positive aspects are highlighted by the verb phrase \textit{is rich in} and verb \textit{help} are used for nutrients and consequences, respectively.

\par \textbf{Explanation Suggestion}: This part represents an alternative behavior that TS4NLE delivers to the user in order to motivate him/her to change his/her lifestyle.
Exploiting the information available within the explanation graph, and possibly collected from both public and private knowledge, TS4NLE generates a \textit{post} suggestion to inform the user about the healthy behavior that he/she can adopt as alternative.
To do that, the data contained in the explanation graph are not sufficient. TS4NLE performs additional meta-reasoning to identify the appropriate content that depends on (i) qualitative properties of the entities involved in the event; (ii) user profile; (iii) other specific violations; (iv) history of messages sent. 
%commented out due to the figs limitation
%The model for generating a suggestion message is shown in Figure~\ref{sugg_message} where, for the sake of readability, we report only the second point of the list: compliance with the user profile.
%\begin{figure}[!htpb]
%\centering
%\includegraphics[width=0.5\textwidth]{figures/sugg_message}
%\caption{TS4NLE model (template and example of violation) for generating text of the suggestion.}
%\label{sugg_message}
%\nat{how is a violation a query? is it constantly queried the system with all possible violations and thus triggered? clarify in the caption, as it is not clear. What is HeLiS? what is an example of query here in this figure, once the violation is detected? it could help explaining these points in the caption with the actual query and actual query syntax. Number order of query and violation and suggestion template, what comes first and in which order the rest follow?}
%\end{figure}
Continuing with the running example, first TS4NLE queries the domain knowledge base through the reasoner to provide a list of alternative foods that are valid alternatives to the violated behavior (e.g., similar-taste relation, list of nutrients, consequences on user health). These alternatives are queried according to some constraints: (i) compliance with the user profile and (ii) compliance with other set up goals. Regarding the first constraint, the reasoner will not return alternative foods that are not appropriate for the specific profile. Let us consider a vegetarian profile: the system does not suggest vegetarian users to consume fish as an alternative to meat, even if fish is an alternative to meat by considering only the nutrients. The second constraint is needed to avoid alternatives that could generate a contradiction with other healthy behavior rules. For example, the system will not propose cheese as alternative to meat if the user has the persuasion goal of cheese reduction.Finally, a control on message history is executed to avoid the suggestion of alternatives recently proposed. %Todo can save space:
Regarding the linguistic aspect, the system uses appropriate verbs, such as \textit{try} or \textit{alternate}, to emphasize the alternative behavior. 
%\nat{moved to be coherent all links to notebooks to the end of each method section}
Both tools\footnote{\url{https://github.com/ivanDonadello/TS4NLE}} and the colaboratory (Colab notebook) session are online\footnote{\url{https://colab.research.google.com/drive/1iCVSt7TFMruSzeg5DswLOzOR1n7xATbz}} for freely creating new use cases %and for testing 
using the TS4NLE approach.

\subsection{TS4NLE suitability analysis: pros and cons}
The use of explanation graphs is an intuitive and effective way for transforming meaningless model outputs into a comprehensive artifact that can be leveraged by targeted users.
Explanation graphs convey formal semantics that: i) can be enriched with other knowledge sources publicly available on the web (e.g. Linked Open Data cloud) or privacy-protected (e.g. user profiles); ii) allow rendering in different formats (e.g. natural language text or audio); and, iii) allow full control over the rendered explanations (i.e. the content of the explanations).
Natural language rendering with a template-system allows full control on the explanations at the price of high effort in domain and user modeling by domain experts. This aspect can be considered the major bottleneck of the TS4NLE approach.
Such bottleneck can be mitigated by using machine learning with human-in-the-loop techniques to increase variability in the generated natural language explanations.

\section{Conclusion}

%In this work, 
We described a number of XAI techniques for extracting explanatory rationales from predictive models that make a particular decision for a particular input. We walked through their implementation so that anyone can adapt them to a specific model and use case, with the ultimate goal of serving a didactic purpose.
%, that can quickly be assessed in terms of their explanations, 
Explanations were aimed at both developer, domain experts and decision maker audiences, and they are open and freely available for instructing and learning\footnote{\url{https://github.com/NataliaDiaz/XAI-guide}}. %Adapting them to his own purpose.
% Future work will address the subject of XAI in neural-symbolic AI with a framework for interactive explainability. \nat{really isnt tackled here?}
Future work should showcase more methods and how they deal with incomplete and multi-modal data \cite{HolzingerEtAl:2021:GraphFusion}. 
%todo methods to explain GNNs and GANs

\section{Acknowledgements}
This research was funded by the French ANRT %(Association Nationale Recherche Technologie - ANRT)
industrial Cifre PhD contracts with SEGULA Technologies and with Tinubu Square. Parts of this work has received funding by the Austrian Science Fund (FWF), Project: P-32554. Díaz-Rodríguez is supported by Juan de la Cierva Incorporación grant IJC2019-039152-I funded by %MCIN/AEI/10.13039/5011%00011033 
MCIN/AEI /10.13039/501100011033 by “ESF Investing in your future” and Google Research Scholar Program. Tulli is supported by the European Union's Horizon 2020 research and innovation programme under grant agreement No 765955 (ANIMATAS Innovative Training Network).

%explain multimodal representations,? and data of other nature. 

% (time series?). Time series are extremely important and often neglected in the medical domain

%ToDo clarify: we used only data where all features are available, not the real data with missing data.
%Tinubu's data doesnt have the name of companies, its anonymized. What matters is the country but is anonymized and encrypted and we can use it,  

%\begin{table}[]
%\begin{tabular}{cc|c|c|}
%\cline{3-4}
%\multicolumn{1}{l}{}                        & \multicolumn{1}{l|}{} & \multicolumn{2}{c|}{Model %Dependence} \\ \cline{3-4} 
%\multicolumn{1}{l}{}                        & \multicolumn{1}{l|}{} & Agnostic      & Specific         %     \\ \hline
%\multicolumn{1}{|c|}{\multirow{3}{*}{Task}} & Image Classification  &               & Grad-CAM (images) %    \\ \cline{2-4} 
%\multicolumn{1}{|c|}{}                      & Other Task            &               &                  %     \\ \cline{2-4} 
%\multicolumn{1}{|c|}{}                      & Other Task            &               &                  %     \\ \hline
%\end{tabular}
%\end{table}

%\section*{References}

%\clearpage
\typeout{}
 \bibliographystyle{plain}
\bibliography{references}

\end{document}